\pdfoutput=1
\pdftrailerid{}
\documentclass[conference]{IEEEtran}
\IEEEoverridecommandlockouts
\usepackage[T1]{fontenc}
\usepackage{newtxtext,newtxmath}
\usepackage{microtype}
\usepackage{amsmath}
\usepackage{graphicx}
\usepackage{tikz}
\usepackage{booktabs}
\usepackage{multirow}
\usepackage{xcolor}
\usepackage{url}
\usepackage[hidelinks]{hyperref}
\hypersetup{pdftitle={LLM Trading Agents Earn Beta, Not Alpha: A Look-Ahead-Free Benchmark},pdfauthor={},pdfcreator={},pdfproducer={}}
\usepackage[capitalise,nameinlink]{cleveref}
\crefname{finding}{Finding}{Findings}
\newcommand{\rih}[1]{\par\smallskip\noindent\textbf{#1}\space\ignorespaces}
\usepackage[most]{tcolorbox}
\definecolor{tradeink}{RGB}{20,66,58}
\definecolor{tradefill}{RGB}{246,249,245}
\definecolor{tradegold}{RGB}{186,134,44}

\newcommand{\findingicon}{\begin{tikzpicture}[baseline=-0.55ex,x=1em,y=1em,line width=0.6pt,line cap=round,line join=round]
\draw[white] (0,-0.28) -- (0.22,-0.02) -- (0.42,-0.16) -- (0.70,0.20);
\fill[white] (0.80,0.32) -- (0.56,0.25) -- (0.73,0.08) -- cycle;
\end{tikzpicture}}
\newtcolorbox[use counter=finding]{finding}[1][]{enhanced, colback=tradefill, colframe=tradeink,
  boxrule=0.4pt, leftrule=2.4pt, arc=1.2mm, outer arc=1.2mm, left=6pt, right=6pt, top=6.5pt, bottom=5pt,
  fontupper=\small, coltitle=white, fonttitle=\footnotesize\bfseries,
  title={\findingicon\enspace\MakeUppercase{Finding}~\thetcbcounter},
  attach boxed title to top left={xshift=5mm, yshift*=-\tcboxedtitleheight/2},
  boxed title style={colback=tradeink, colframe=tradegold, boxrule=0.5pt, arc=0.8mm, left=4pt, right=5pt, top=1.2pt, bottom=1.2pt},
  overlay={\draw[tradegold, line width=1.4pt] ([xshift=-16mm]frame.south east) -- ([xshift=-2.5mm]frame.south east);},
  before skip=10pt, after skip=8pt, #1}

\newcommand{\bench}{rta-bench-1}

\newcommand{\abBaseNewsShare}{29}

\newcommand{\abCalNewsShare}{\ensuremath{-}1}

\newcommand{\abCalRet}{\ensuremath{-}1.4}
\newcommand{\abCalRetHi}{+1.9}
\newcommand{\abCalRetLo}{\ensuremath{-}4.4}

\newcommand{\abNoNewsDec}{+0.09}
\newcommand{\abNoNewsDecHi}{+0.24}
\newcommand{\abNoNewsDecLo}{\ensuremath{-}0.08}

\newcommand{\abNoNewsRet}{\ensuremath{-}0.3}
\newcommand{\abNoNewsRetHi}{+2.7}
\newcommand{\abNoNewsRetLo}{\ensuremath{-}3.4}

\newcommand{\abShufDec}{+0.15}
\newcommand{\abShufDecHi}{+0.38}
\newcommand{\abShufDecLo}{\ensuremath{-}0.05}

\newcommand{\abShufRet}{\ensuremath{-}2.2}
\newcommand{\abShufRetHi}{+1.1}
\newcommand{\abShufRetLo}{\ensuremath{-}5.3}

\newcommand{\alphaCells}{18}
\newcommand{\alphaHiPct}{0.1}
\newcommand{\alphaLoPct}{\ensuremath{-}4.7}
\newcommand{\alphaMaxAbsT}{1.79}
\newcommand{\alphaMeanPct}{\ensuremath{-}2.3}
\newcommand{\alphaSeMedianPct}{7.2}

\newcommand{\benchCash}{100,000}
\newcommand{\benchDecideEvery}{5}
\newcommand{\benchMaxNews}{3}
\newcommand{\benchMaxPositions}{10}
\newcommand{\benchSectors}{11}
\newcommand{\benchStocks}{56}
\newcommand{\betaMktHi}{0.90}
\newcommand{\betaMktLo}{0.01}
\newcommand{\chkCleanTotal}{24}
\newcommand{\chkDecCaught}{48}

\newcommand{\chkRuns}{12}

\newcommand{\cleanRuns}{109}
\newcommand{\daysTwenty}{253}
\newcommand{\daysTwentySix}{217}
\newcommand{\daysTwentyTwo}{251}

\newcommand{\ddSpyTwenty}{33.7}
\newcommand{\ddSpyTwentySix}{8.9}
\newcommand{\ddSpyTwentyTwo}{24.5}

\newcommand{\decBuyMeanPct}{0.18}

\newcommand{\decBuyT}{1.3}
\newcommand{\decHiPct}{0.09}
\newcommand{\decHit}{50}
\newcommand{\decLoPct}{\ensuremath{-}0.30}
\newcommand{\decMeanPct}{\ensuremath{-}0.10}

\newcommand{\decSellAbsPct}{0.32}

\newcommand{\decSellStockT}{2.2}

\newcommand{\egBuyEg}{7.9}
\newcommand{\egBuyNone}{1.6}
\newcommand{\egBuyOther}{3.3}
\newcommand{\egModels}{five}

\newcommand{\egRatioHi}{2.8}
\newcommand{\egRatioLo}{1.5}
\newcommand{\egRetEg}{0.31}
\newcommand{\egRetEgT}{0.9}
\newcommand{\egRetNone}{0.12}
\newcommand{\egRetNoneT}{1.4}
\newcommand{\egRetOther}{0.13}
\newcommand{\egRetOtherT}{1.6}
\newcommand{\entryCount}{26}
\newcommand{\evAnalystAfterClosePct}{1}

\newcommand{\evCalSig}{5}

\newcommand{\evCells}{22}
\newcommand{\evEarnAfterClosePct}{39}
\newcommand{\evEarnNegCal}{+0.79}

\newcommand{\evEarnNegCalT}{3.0}
\newcommand{\evEarnNegPub}{+0.08}

\newcommand{\evEarnNegPubT}{0.6}

\newcommand{\evEarnPosPubFive}{+0.34}
\newcommand{\evEarnPosPubFiveT}{2.2}

\newcommand{\evGapCount}{12}
\newcommand{\evGapTotal}{12}
\newcommand{\evGuideAfterClosePct}{38}
\newcommand{\evGuideNegCal}{+0.53}

\newcommand{\evGuideNegCalT}{2.7}

\newcommand{\evGuidePosCal}{+0.52}

\newcommand{\evGuidePosCalT}{2.9}
\newcommand{\evGuidePosPub}{+0.15}

\newcommand{\evGuidePosPubTT}{1.98}

\newcommand{\evSectorPosPubFive}{+0.51}
\newcommand{\evSectorPosPubFiveT}{2.1}
\newcommand{\expoJuneTwenty}{42}
\newcommand{\expoLowTwenty}{14}
\newcommand{\expoMaxAll}{98}

\newcommand{\expoMinAll}{2}

\newcommand{\expoPeakTwenty}{69}
\newcommand{\expoTwenty}{49}
\newcommand{\expoTwentySix}{57}
\newcommand{\expoTwentyTwo}{44}
\newcommand{\exposureShareMedian}{78}

\newcommand{\firstDayTwentySix}{January~2}

\newcommand{\highCells}{9}
\newcommand{\highNeg}{9}

\newcommand{\horBuyTwenty}{+1.01}
\newcommand{\horBuyTwentyT}{2.8}

\newcommand{\horSellTwenty}{+0.94}
\newcommand{\horSellTwentyT}{2.6}

\newcommand{\icLlm}{0.017}
\newcommand{\icLlmT}{1.26}

\newcommand{\icNews}{0.004}
\newcommand{\icNewsT}{0.28}

\newcommand{\icVol}{0.015}
\newcommand{\icVolT}{1.28}

\newcommand{\labMissingPct}{0.4}

\newcommand{\lastDayTwentySix}{November~11}

\newcommand{\latAllEarnEvents}{382}

\newcommand{\latAllEarnWhole}{+0.88}

\newcommand{\latAllEarnWholeT}{3.2}
\newcommand{\latAllEvents}{3,289}

\newcommand{\latAllOtherRegOpen}{\ensuremath{-}0.14}

\newcommand{\latAllOtherRegOpenT}{\ensuremath{-}2.7}

\newcommand{\latCmpAllOnly}{two}

\newcommand{\latCmpCells}{24}

\newcommand{\latCmpSame}{21}
\newcommand{\latEarnEvents}{159}
\newcommand{\latEarnFirst}{+0.40}

\newcommand{\latEarnFirstT}{2.03}
\newcommand{\latEarnLeft}{+0.41}

\newcommand{\latEarnLeftT}{1.1}
\newcommand{\latEarnOpen}{+0.31}

\newcommand{\latEarnOpenT}{1.0}
\newcommand{\latEarnOutFirst}{+0.68}

\newcommand{\latEarnOutFirstT}{2.4}

\newcommand{\latEarnOutWhole}{+1.64}

\newcommand{\latEarnOutWholeT}{2.6}
\newcommand{\latEarnOutsidePct}{66}

\newcommand{\latEarnRegWhole}{\ensuremath{-}0.31}

\newcommand{\latEarnShareTen}{38}
\newcommand{\latEarnWhole}{+0.92}

\newcommand{\latEarnWholeT}{1.9}
\newcommand{\latEvents}{1,079}

\newcommand{\latFiveDays}{58}
\newcommand{\latFiveEarnEvents}{38}

\newcommand{\latFiveEarnZero}{+0.28}

\newcommand{\latFiveEarnZeroT}{0.6}
\newcommand{\latFiveOtherEvents}{231}

\newcommand{\latFiveOtherZero}{+0.07}

\newcommand{\latFiveOtherZeroT}{0.3}

\newcommand{\latMixed}{734}

\newcommand{\latOneDays}{15}
\newcommand{\latOneEarnEvents}{11}

\newcommand{\latOneOtherEvents}{56}

\newcommand{\latOtherFirst}{+0.09}

\newcommand{\latOtherFirstT}{3.0}
\newcommand{\latOtherLeft}{+0.17}

\newcommand{\latOtherLeftT}{1.8}
\newcommand{\latOtherOpen}{+0.04}

\newcommand{\latOtherOpenT}{0.6}

\newcommand{\latOtherRegWhole}{+0.20}

\newcommand{\latOtherWhole}{+0.36}

\newcommand{\latOtherWholeT}{3.5}
\newcommand{\latPreEarnFive}{+0.04}
\newcommand{\latPreEarnFiveHour}{\ensuremath{-}0.09}

\newcommand{\latPreEarnFiveHourT}{\ensuremath{-}1.0}

\newcommand{\latPreEarnFiveT}{1.0}
\newcommand{\latPreEarnHour}{+0.08}

\newcommand{\latPreEarnHourT}{1.8}

\newcommand{\latPreOtherFive}{0.00}
\newcommand{\latPreOtherFiveHour}{+0.03}

\newcommand{\latPreOtherFiveHourT}{0.5}

\newcommand{\latPreOtherFiveT}{0.1}
\newcommand{\latPreOtherHour}{+0.04}

\newcommand{\latPreOtherHourT}{2.1}

\newcommand{\latStocks}{ten}
\newcommand{\lexIcCal}{0.032}
\newcommand{\lexIcCalT}{3.4}
\newcommand{\lexIcPub}{0.005}
\newcommand{\lexIcPubT}{0.6}
\newcommand{\llamaDecMax}{20}
\newcommand{\llamaDecMin}{1}
\newcommand{\llmBreakEvenBp}{2.7}
\newcommand{\llmIcCal}{0.050}
\newcommand{\llmIcCalT}{5.6}
\newcommand{\llmIcPub}{0.021}
\newcommand{\llmIcPubT}{2.5}
\newcommand{\llmTurnoverPct}{82}
\newcommand{\memDirBaseBefore}{53}
\newcommand{\memDirBefore}{75}
\newcommand{\memHaikuRecall}{0.24}
\newcommand{\memHaikuRecallHi}{0.54}
\newcommand{\memHaikuRecallLo}{\ensuremath{-}0.09}
\newcommand{\memLlamaRecall}{0.06}

\newcommand{\memRecallAfter}{0.25}
\newcommand{\memRecallAfterHi}{0.55}
\newcommand{\memRecallAfterLo}{\ensuremath{-}0.05}
\newcommand{\memRecallBefore}{0.69}
\newcommand{\memRecallBeforeHi}{0.80}
\newcommand{\memRecallBeforeLo}{0.51}
\newcommand{\memScore}{\ensuremath{-}0.047}
\newcommand{\memSonnetRecall}{0.52}
\newcommand{\memSonnetRecallHi}{0.72}
\newcommand{\memSonnetRecallLo}{0.26}
\newcommand{\modelSpreadTwentySix}{8.7}
\newcommand{\nDecisions}{9,101}
\newcommand{\nModels}{six}
\newcommand{\nModelsFit}{five}
\newcommand{\nNamedPairs}{43}
\newcommand{\nRuns}{91}

\newcommand{\nSteps}{4,394}
\newcommand{\namedHigherBeta}{38}
\newcommand{\namedMinusBlindBeta}{0.21}
\newcommand{\namedMinusBlindRet}{1.3}

\newcommand{\newsScoreBase}{3.0}
\newcommand{\newsScoreCells}{9}
\newcommand{\newsScoreCoefHi}{4.4}
\newcommand{\newsScoreCoefLo}{0.3}

\newcommand{\newsScoreTHi}{7.8}
\newcommand{\newsScoreTLo}{2.2}
\newcommand{\nfAlphaMaxAbsT}{1.91}

\newcommand{\nfRTwoFour}{0.66}
\newcommand{\nfRTwoSeven}{0.66}
\newcommand{\nfRuns}{91}
\newcommand{\nfScoreNeg}{1}
\newcommand{\nfScorePos}{10}
\newcommand{\nfToneNeg}{7}
\newcommand{\nfTonePos}{5}

\newcommand{\noReplyRuns}{4}
\newcommand{\noReplySteps}{4}
\newcommand{\nullDecBottom}{6}
\newcommand{\nullDecPct}{54}
\newcommand{\nullDecTop}{7}
\newcommand{\nullPortAnyPct}{55}
\newcommand{\nullPortSectorPct}{48}
\newcommand{\nullRuns}{91}
\newcommand{\pilotExpoBase}{44}
\newcommand{\pilotExpoRisk}{19}
\newcommand{\pilotExpoTrend}{15}
\newcommand{\pilotHitHi}{42}
\newcommand{\pilotHitLo}{37}
\newcommand{\pilotRiskHi}{14.5}
\newcommand{\pilotRiskLo}{\ensuremath{-}0.7}
\newcommand{\pilotRiskRet}{7.3}
\newcommand{\pilotRiskSplitLevel}{4.6}
\newcommand{\pilotRiskSplitOther}{\ensuremath{-}0.3}
\newcommand{\pilotRiskSplitSelection}{1.0}
\newcommand{\pilotRiskSplitTiming}{3.7}
\newcommand{\pilotRiskTotal}{9.0}
\newcommand{\pilotRiskWins}{8}
\newcommand{\pilotTrendRet}{8.9}
\newcommand{\pilotTrendSplitLevel}{6.5}
\newcommand{\pilotTrendSplitOther}{+0.3}
\newcommand{\pilotTrendSplitSelection}{\ensuremath{-}2.3}
\newcommand{\pilotTrendSplitTiming}{6.0}
\newcommand{\pilotTrendTotal}{10.5}
\newcommand{\plAboveExpo}{\ensuremath{-}3.8}

\newcommand{\plCapExpo}{\ensuremath{-}7.6}

\newcommand{\plCapSel}{\ensuremath{-}5.2}

\newcommand{\plCashFiftyExpoAbs}{49.6}

\newcommand{\plCashFiftyMddAbs}{18.4}

\newcommand{\plCashFiftyRet}{+0.1}

\newcommand{\plCashFiftyRetTwenty}{\ensuremath{-}24.8}

\newcommand{\plCashFiftyRetTwentyTwo}{+24.3}

\newcommand{\plCashFiftySel}{\ensuremath{-}0.5}

\newcommand{\plCashThirtyExpoAbs}{36.7}

\newcommand{\plCashThirtyMddAbs}{12.7}

\newcommand{\plCashThirtyRet}{\ensuremath{-}4.2}

\newcommand{\plCashThirtyRetTwenty}{\ensuremath{-}21.4}

\newcommand{\plCashThirtyRetTwentyTwo}{+11.9}

\newcommand{\plCashThirtySel}{\ensuremath{-}1.5}

\newcommand{\plNewsSixMddAbs}{4.5}
\newcommand{\plNewsSixMddDown}{five}
\newcommand{\plNewsSixMddHi}{\ensuremath{-}0.2}
\newcommand{\plNewsSixMddLo}{\ensuremath{-}9.9}

\newcommand{\plNewsSixVolAbs}{3.5}

\newcommand{\plNewsSixVolHi}{\ensuremath{-}0.2}
\newcommand{\plNewsSixVolLo}{\ensuremath{-}8.5}

\newcommand{\plNewsZeroExpoAbs}{4.3}

\newcommand{\plNewsZeroExpoHi}{\ensuremath{-}1.6}

\newcommand{\plNewsZeroExpoLo}{\ensuremath{-}6.9}

\newcommand{\plPairs}{six}

\newcommand{\plRegimeExpoAbs}{30.3}

\newcommand{\plRegimeMddAbs}{9.4}

\newcommand{\plRegimeVolAbs}{10.8}

\newcommand{\plRetMinP}{0.19}

\newcommand{\plRiskExpoAbs}{28.1}

\newcommand{\plRiskLevel}{\ensuremath{-}1.1}

\newcommand{\plRiskMddAbs}{8.0}

\newcommand{\plRiskRet}{\ensuremath{-}7.5}

\newcommand{\plRiskRetTwenty}{\ensuremath{-}25.5}

\newcommand{\plRiskRetTwentyTwo}{+12.6}

\newcommand{\plRiskSel}{\ensuremath{-}3.4}

\newcommand{\plRiskTiming}{\ensuremath{-}2.2}

\newcommand{\plRiskTotal}{\ensuremath{-}6.5}

\newcommand{\plRiskVolAbs}{10.1}

\newcommand{\plSeeds}{two}

\newcommand{\plStopExpoAbs}{10.4}

\newcommand{\plStopSel}{\ensuremath{-}5.8}

\newcommand{\plStopWorstAbs}{2.3}

\newcommand{\rTwoMedian}{0.66}
\newcommand{\relVolPosSig}{6}

\newcommand{\replayRuns}{91}
\newcommand{\rerunAbsDiff}{5.4}
\newcommand{\rerunMaxAbsT}{1.79}
\newcommand{\rerunPairs}{24}
\newcommand{\rerunRuns}{24}

\newcommand{\retEwTwenty}{+40.1}
\newcommand{\retEwTwentySix}{+11.4}
\newcommand{\retEwTwentyTwo}{\ensuremath{-}12.4}
\newcommand{\retHighTwentySix}{8.1}
\newcommand{\retLowTwentySix}{\ensuremath{-}0.6}

\newcommand{\retSpyTwenty}{+17.2}
\newcommand{\retSpyTwentySix}{+12.1}
\newcommand{\retSpyTwentyTwo}{\ensuremath{-}18.6}
\newcommand{\retTrendTwenty}{+67.4}
\newcommand{\retTrendTwentySix}{+19.0}

\newcommand{\riskBadTwenty}{+1.09}

\newcommand{\riskBadTwentyT}{2.4}

\newcommand{\riskCalExpo}{+0.7}

\newcommand{\riskCalMdd}{+2.0}
\newcommand{\riskCalMddHi}{+3.4}
\newcommand{\riskCalMddLo}{+0.4}

\newcommand{\riskCalPairs}{18}

\newcommand{\riskCalWorst}{+1.1}
\newcommand{\riskCalWorstHi}{+1.8}
\newcommand{\riskCalWorstLo}{+0.4}

\newcommand{\riskCalWorstPos}{17}
\newcommand{\riskCrashPairs}{six}
\newcommand{\riskCrashWith}{16.6}
\newcommand{\riskCrashWithout}{15.9}

\newcommand{\riskKeptTwenty}{+1.23}
\newcommand{\riskLowWith}{11}
\newcommand{\riskLowWithout}{8}

\newcommand{\riskNoNewsCrashHi}{+2.3}
\newcommand{\riskNoNewsCrashLo}{\ensuremath{-}3.2}

\newcommand{\riskNoNewsExpo}{+3.8}

\newcommand{\riskNoNewsExpoGeminiAbs}{6.4}
\newcommand{\riskNoNewsExpoHi}{+9.6}
\newcommand{\riskNoNewsExpoLo}{\ensuremath{-}2.2}

\newcommand{\riskNoNewsExpoLunaAbs}{13.9}

\newcommand{\riskNoNewsPairs}{18}

\newcommand{\riskNoNewsWorst}{+0.8}
\newcommand{\riskNoNewsWorstHi}{+1.5}
\newcommand{\riskNoNewsWorstLo}{+0.2}

\newcommand{\riskNoNewsWorstPos}{12}
\newcommand{\riskPeakWith}{90}
\newcommand{\riskPeakWithout}{95}
\newcommand{\riskSellBadNews}{31.4}
\newcommand{\riskSellBadNoNews}{31.2}
\newcommand{\riskSellDiff}{+1.3}
\newcommand{\riskSellDiffHi}{+4.5}
\newcommand{\riskSellDiffLo}{\ensuremath{-}1.7}

\newcommand{\riskSellNewsCells}{9}

\newcommand{\riskSellNewsNeg}{6}

\newcommand{\riskSellOtherNews}{20.3}
\newcommand{\riskSellOtherNoNews}{23.7}

\newcommand{\riskSoldTwenty}{+1.64}
\newcommand{\seedAbsDiff}{5.2}
\newcommand{\seedPairs}{54}
\newcommand{\sharpeCIWidth}{3.9}
\newcommand{\smAvgGross}{47}
\newcommand{\smLsTwenty}{\ensuremath{-}14.4}
\newcommand{\smLsTwentyTwo}{9.0}
\newcommand{\smMaxGross}{125}
\newcommand{\smRuns}{54}

\newcommand{\smShortsTwentyAbs}{2.2}
\newcommand{\smShortsTwentyStockT}{3.1}

\newcommand{\smShortsTwentyTwo}{1.1}
\newcommand{\smShortsTwentyTwoStockT}{\ensuremath{-}3.2}

\newcommand{\timingAfterCalBp}{103.0}
\newcommand{\timingAfterClose}{178}
\newcommand{\timingAfterPubBp}{\ensuremath{-}8.3}
\newcommand{\timingCalBp}{32.0}

\newcommand{\timingCalTrades}{444}

\newcommand{\timingPubBp}{\ensuremath{-}6.7}

\newcommand{\timingPubTrades}{458}
\newcommand{\ttCalAfterClose}{+0.53}

\newcommand{\ttCalAfterCloseT}{1.5}
\newcommand{\ttCalAfterOpen}{0.00}

\newcommand{\ttCalAfterOpenT}{0.0}

\newcommand{\ttCalAllOpen}{+0.02}

\newcommand{\ttCalAllOpenT}{0.2}

\newcommand{\ttPubAllOpen}{\ensuremath{-}0.04}

\begin{document}

\title{LLM Trading Agents Earn Beta, Not Alpha: A Look-Ahead-Free Benchmark}

\author{\IEEEauthorblockN{\textit{Jackson Wang}}}

\date{}
\maketitle
\thispagestyle{plain}
\pagestyle{plain}

\begin{abstract}
Large language model (LLM) trading agents now trade stocks for their users, and benchmarks rank models by their agents' returns. Yet a single run's return says little about skill. In our study, two runs of the same model that differed only in the random seed ended \seedAbsDiff~points apart on average, so telling skill from luck takes many backtests across markets, models and seeds. However, live arenas trade each day only once, and backtests either use the few months after a model's training cutoff or run on years the model may remember, often with news joined to prices by calendar date. In this paper, we present \bench, a backtest over three US markets (a rally, a bear market and a crash). It shows each headline only after its publication minute, plays each period blind and named, and checks every run for look-ahead. It also splits each return into factor exposure and alpha and scores every decision against random choices of the same day and sector. In \nRuns~runs of \nModels~models from four vendors, no alpha was significant, market and style exposure accounted for a median \exposureShareMedian\% of the return, and the decisions did no better than random stocks of the same sector. The agents bought the stocks with the most positive news, yet in ablations they earned as much without their headlines. Still, the agents carried out rules reliably, since a single cash rule in the prompt cut a GPT-based agent's share in stocks by \plRegimeExpoAbs~points in all \plPairs~pairs. Builders should therefore supply the edge themselves and report exposures and alpha, and benchmarks should replay news by publication time.
\end{abstract}

\section{Introduction}
\label{sec:intro}

Large language model (LLM) trading agents now read prices and news and decide what to buy and sell. Research systems give the model a layered memory, a team of analyst agents or a reinforcement-learning objective~\cite{yu2025finmem,xiao2024tradingagents,yu2024fincon,xiao2025tradingr1}, thousands of agents funded by their users already trade in production~\cite{barton2026trading}, and benchmarks rank models by the returns their agents make in replayed or live markets~\cite{chen2025stockbench,yu2025livetradebench,fan2025aitrader,qian2026agents}. Many of these results beat the market. However, a single run's return says little about skill. In our runs, two seeds of the same model ended \seedAbsDiff~points of return apart on average, more than half the \modelSpreadTwentySix~points between the best and the worst model of 2026. A return can also come from news seen before its publication, from a model's memory of the period, or from simply holding a rising market. Telling skill from luck therefore takes many backtests, over markets that rose and fell, with several models and seeds, and a way to separate what the agent picked from what the market gave it.

Existing evaluations cannot run such backtests. Live arenas avoid leaks by trading the real market~\cite{yu2025livetradebench,fan2025aitrader,qian2026agents,li2025time}, but each day happens once, so LiveTradeBench's 50-day test cannot be repeated or rerun for a new model~\cite{yu2025livetradebench}. Backtests can be repeated, but they trade length against memory. StockBench avoids contamination by trading only the months after the models' training cutoffs~\cite{chen2025stockbench}, which leaves a short sample from one market. FinSABER backtests LLM strategies over two decades~\cite{li2026financial}, but those are years the models have read about, and pretrained models recall market data from before their cutoff~\cite{lopezlira2025memorization,gao2025detecting}. Hiding tickers, as BlindTrade does, lowers this recall but does not measure it~\cite{jeon2026blindfolded}. Backtests also often join news to prices by calendar date, so an agent that decides at a day's close also receives the headlines published that evening. Two recent benchmarks go further. KTD-Fin masks identities and decomposes returns into market, style and selection on Chinese stocks~\cite{zhu2026knowing}, and OpenPM requires every record given to an agent to have existed at decision time~\cite{cai2026openpm}. Neither measures how much the timing of news inflates what news seems to be worth, and neither checks the agents' single decisions against what random choices would have earned on the same days.

In this paper, we present \bench~\cite{rta2026github}, a backtest of LLM trading agents over three periods of the US stock market, one that rose, one that fell slowly and one that crashed. Building it meant solving three technical challenges. The first is the timing of news, which alone can create an edge. For example, a rule that buys after an earnings beat or a raised outlook and sells short after a miss or a cut earns \timingCalBp~basis points per trade on our 2026 data when headlines are joined by calendar date and traded from close to close. It earns \timingPubBp~bp once they are replayed by publication time and traded from the next open, because \timingAfterClose~of the headlines came out after the close (\cref{fig:timing}). \bench{} therefore shows each headline only after its publication minute and fills orders at the next open, with costs. The second challenge is memory, because a long backtest has to use years before a model's training cutoff. \bench{} therefore plays each period both blind and with real names, splits each model's scores at its cutoff, and asks the model directly what it remembers. The third challenge is checking the result, because a replay can still leak the future through its code, and a return alone does not show where it came from. \bench{} tests every run for look-ahead by replaying the model's own replies on markets whose future is random, splits each return into factor exposure and alpha, and scores each decision against random choices of the same day and sector. Unlike KTD-Fin and OpenPM, \bench{} also measures what the timing of news is worth and whether single decisions beat chance.

We ran \nModels~models from four vendors through the benchmark, from a small open-weight Llama model to mid-size commercial models, in \nRuns~runs. The agents earned what their exposures earned. Not one run had a significant four-factor alpha. In the median run, market and style exposure accounted for \exposureShareMedian\% of the return. Their single decisions did no better than chance either. Against the same decision with a random stock of the same sector on the same day, a run's mean decision had a percentile rank of \nullDecPct{} on average, where no skill gives 50. To find out what the agents trade on, we recovered the exact prompt of every decision and regressed each model's choices on the features in that prompt. All \nModelsFit~models that traded more than a few times bought the stocks with the highest news scores, yet the news score had a rank correlation of \icNews{} with the next five days' return over SPY (\cref{sec:returns}). In ablations with two models, the agents earned as much without the headlines, or with another stock's, and their decisions were no worse (\cref{sec:news}). In the intraday prices of 2026, what was left of a headline's move an hour after its publication was not measurably above zero (\cref{sec:latency}).

These results change what an LLM agent's backtest return should be taken to mean. Because prompt rules and account rules changed how much stock the agents held rather than which stocks they picked (\cref{sec:robustness}), a builder who reports only returns cannot tell a better agent from a more invested one. The agents are still useful, but as executors of a builder's rules rather than as a source of edge. For example, one prompt rule that asks for 30\% cash in uncertain or risk-off markets cut GPT-5.6 Luna's share in stocks by \plRegimeExpoAbs~points in all \plPairs~pairs and its maximum drawdown by \plRegimeMddAbs~points (\cref{sec:prompts2}). Agent builders should therefore supply the edge from data or signals of their own, report each agent's factor exposures and alpha, and score its decisions against random choices of the same day and sector. Benchmarks should replay news by publication time, check the replay on the days the agent decides, and run several seeds. We make three contributions:

\begin{itemize}
\item \textbf{A Benchmark That Keeps Tomorrow's News Out.} We present \bench, which replays three US markets with each headline shown only after its publication minute and scores each run's exposures, alpha and decisions against random choices. Its look-ahead check, with cuts on decision days, caught all \chkDecCaught~leaks that we planted, while fixed cut days missed every one-day leak in 2026 (\cref{sec:benchmark,sec:checkresults}).
\item \textbf{Twelve Findings, Nine Against Common Expectations.} We report twelve findings, each in a box next to the result that supports it, with its control and its count. Nine of them run against what agent builders and earlier benchmarks commonly expect (\cref{tab:findings}). For example, a trend-following rule made more money than every agent, and the agents earned as much without their headlines (\cref{sec:returns,sec:news,sec:robustness}).
\item \textbf{Recommendations and an Open Benchmark.} We turn the findings into recommendations for agent builders, benchmark designers and people who fund trading agents, each tied to the finding behind it (\cref{sec:recommendations}). The package that runs every check, the frozen markets and the rta-bench-1 leaderboard are open source~\cite{rta2026github}.
\end{itemize}

\section{Testing LLM Trading Agents}
\label{sec:background}

\label{sec:agents}An LLM trading agent is a program that asks a large language model what to buy and sell. At each decision time it collects prices, indicators and news into a prompt, asks the model for orders, and sends those orders to a broker or a simulator. Published agents add more machinery around the model. FinMem gives the model a layered memory of past events~\cite{yu2025finmem}, and FinAgent adds tools and chart reading~\cite{zhang2024multimodal}. FinCon and TradingAgents split the work among several agents in roles such as analyst, trader and risk manager~\cite{yu2024fincon,xiao2024tradingagents}. Trading-R1 goes further and trains the model itself with reinforcement learning~\cite{xiao2025tradingr1}. These systems are usually judged by the return and the Sharpe ratio of a backtest over a few months, and many of them report results above the market.

Benchmarks for these agents come in two kinds. Backtest benchmarks replay a stretch of history and let every agent trade it. InvestorBench covers stocks, cryptocurrencies and exchange-traded funds~\cite{li2024investorbench}, StockBench lets agents trade stocks for several months on daily prices, fundamentals and news~\cite{chen2025stockbench}, and FinSABER extends backtests to two decades and more than 100 symbols~\cite{li2026financial}. Live arenas trade the present instead. LiveTradeBench, AI-Trader and Agent Market Arena stream prices and news as they happen and let models trade them~\cite{yu2025livetradebench,fan2025aitrader,qian2026agents}, and DeepFund runs a live fund for the same reason~\cite{li2025time}. A live arena cannot leak the future, because the future has not happened yet. However, each trading day happens once. Two agents can only be compared if they ran on the same days, and a result cannot be run again. A model released next year can never trade the days that the current leaderboard was built on.

\label{sec:pitfalls}A backtest can be run again and compared, but its return mixes skill with four other things. Each of them can make an agent look better than it is, and none of them shows up in the return itself.

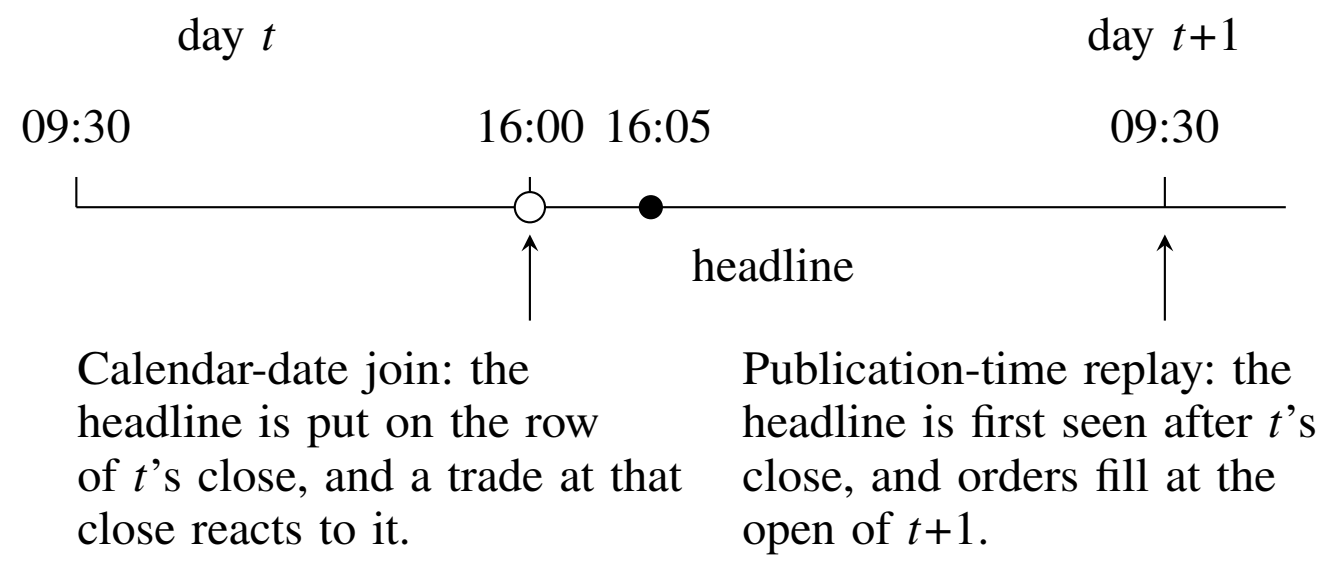
\begin{figure}[t]
\centering
\begin{tikzpicture}[x=1mm,y=1mm,font=\small,>=stealth]
\draw (2,30) -- (82,30);
\foreach \x/\l in {2/09:30,32/16:00,74/09:30} {\draw (\x,30) -- (\x,32); \node[anchor=south] at (\x,33) {\l};}
\draw[fill=white] (32,30) circle (1);
\fill (40,30) circle (0.8);
\node[anchor=south] at (40.5,33) {16:05};
\node[anchor=south] at (12,38.5) {day $t$};
\node[anchor=south] at (74,38.5) {day $t{+}1$};
\node[anchor=north west] at (41.5,28.5) {headline};
\draw[->] (32,22.5) -- (32,28.2);
\draw[->] (74,22.5) -- (74,28.2);
\node[anchor=north west,text width=40mm,align=left,inner sep=0pt] at (2,20.5) {Calendar-date join: the headline is put on the row of $t$'s close, and a trade at that close reacts to it.};
\node[anchor=north west,text width=38mm,align=left,inner sep=0pt] at (46,20.5) {Publication-time replay: the headline is first seen after $t$'s close, and orders fill at the open of $t{+}1$.};
\end{tikzpicture}
\caption{A headline published five minutes after the close. Joined by calendar date, it is put on the row of a close that came before it, and a close-to-close backtest trades on it. Replayed by publication time, it is first visible after that close and can only be traded at the next open. On the 2026 data, this difference turns a headline rule that earns \timingPubBp~bp per trade into one that earns \timingCalBp~bp.}
\label{fig:timing}
\end{figure}

\rih{News Read Before It Was Published.} A backtest must show the agent only the news that existed at the moment of the decision. Many news backtests instead join headlines to prices by calendar date, so a headline published at 16:05 New York time is put on the row of that day's 16:00 close (\cref{fig:timing}). A strategy that trades at that close then reacts to news that did not yet exist when the price was set. The effect is large even for a simple rule. We took one headline rule on the 2026 data: buy for one day after a headline says a company beat estimates or raised its outlook, and sell short after a miss or a cut. Joined by calendar date and traded close to close, the rule earned \timingCalBp~basis points (bp) per trade over \timingCalTrades~trades. Joined by publication time and traded from the next open, the same rule earned \timingPubBp~bp per trade over \timingPubTrades~trades. The difference is driven by the \timingAfterClose~headlines that were published after the close. By calendar date they earned \timingAfterCalBp~bp per trade, and by publication time they earned \timingAfterPubBp~bp.

\rih{A Model That Remembers the Test Period.} Even with the right timestamps, the model may know what happened next. Language models recall economic and market data from before their training cutoff~\cite{lopezlira2025memorization,gao2025detecting}, and they carry knowledge of what followed a historical date into decisions made at that date~\cite{jia2026hindsightbench,glasserman2023assessing}. We asked GPT-5.6 Luna, with no data at all, how individual stocks moved over 20-day windows of 2022. Its guesses ranked the actual moves with a rank correlation of \memRecallBefore{} (95\% interval \memRecallBeforeLo{} to \memRecallBeforeHi{}). It named the direction correctly \memDirBefore\% of the time against \memDirBaseBefore\% for always guessing the more common direction. After its training cutoff the same correlation fell to \memRecallAfter{} (\memRecallAfterLo{} to \memRecallAfterHi{}).

\rih{Exposure Taken for Skill.} A portfolio that holds stocks earns the market's return, and a portfolio that leans toward growth or recent winners earns those styles' returns. Asset pricing separates these exposures from stock selection with factor models~\cite{fama1993common,carhart1997persistence,jegadeesh1993returns}. Without that step, a bull market makes every long agent look skilled. The benchmark's own stock list shows the size of the effect. The \benchStocks~large caps, held in equal weights, returned \retEwTwenty\% in 2020 against \retSpyTwenty\% for the S\&P 500 exchange-traded fund (SPY), because these are the companies that went on to stay large. KTD-Fin found on Chinese stocks that most of what LLM agents earn is market and style exposure~\cite{zhu2026knowing}.

\rih{One Window, One Run.} A year of daily returns is a short sample. Across the \entryCount~leaderboard entries we started from, the 95\% bootstrap interval of a run's Sharpe ratio was \sharpeCIWidth~units wide at the median~\cite{lo2002statistics}. A benchmark that tries many agents and prompts will find some that look good by chance~\cite{bailey2014deflated,harvey2015cross,white2000reality}. Language models also sample their answers, so the same agent on the same days can trade differently from run to run.

\section{The Benchmark}
\label{sec:benchmark}

\begin{figure*}[t]
\centering
\begin{tikzpicture}[x=1cm,y=1cm,font=\small,>=stealth,
  box/.style={draw,rounded corners=1pt,align=center,inner sep=3pt,fill=white},
  data/.style={box,fill=black!8},
  num/.style={inner sep=1pt,font=\footnotesize\bfseries}]
\node[data,text width=29mm] (px) at (1.6,0.8) {Prices of \benchStocks~stocks,\\ index funds and T-bills};
\node[data,text width=29mm] (nw) at (1.6,-0.8) {Headlines with the\\ minute of publication};
\node[box,text width=36mm] (obs) at (5.85,0) {Prompt at the close:\\ prices, indicators and account,\\ up to \benchMaxNews~headlines per stock\\ published before the close};
\node[box,text width=19mm] (llm) at (9.5,0) {LLM agent\\ replies with\\ buys and sells};
\node[box,text width=24mm] (fill) at (12.5,0) {Orders fill at the\\ next open, with\\ trading costs};
\node[box,text width=21mm] (rec) at (15.75,0) {Recorded run:\\ replies, fills,\\ daily equity};
\node[font=\footnotesize,anchor=south] at (obs.north) {blind: aliases, day numbers, rebased prices};
\draw[->] (px.east) -- (obs.west|-px.east);
\draw[->] (nw.east) -- (obs.west|-nw.east);
\draw[->] (obs) -- node[num,above=3pt]{(1)} (llm);
\draw[->] (llm) -- node[num,above=3pt]{(2)} (fill);
\draw[->] (fill) -- node[num,above=3pt]{(3)} (rec);
\draw[->] (fill.south) -- ++(0,-0.45) -| node[pos=0.25,below,font=\footnotesize]{account carried to the next decision, \benchDecideEvery~closes later} (obs.south);
\node[box,text width=38mm] (chk) at (2.3,-3.35) {Look-ahead check: replay the\\ replies on markets with a\\ random future after each cut};
\node[box,text width=38mm] (fac) at (6.75,-3.35) {Exposure and alpha:\\ four-factor regression\\ of the daily returns};
\node[box,text width=38mm] (nul) at (11.2,-3.35) {Decisions against random\\ stocks of the same day,\\ direction and sector};
\node[box,text width=38mm] (pol) at (15.65,-3.35) {Features traded: the\\ prompt features that\\ predict a buy};
\draw (rec.south) -- (15.75,-2.05);
\draw (2.3,-2.05) -- node[num,pos=0.97,above=3pt]{(4)} (15.75,-2.05);
\foreach \n in {chk,fac,nul,pol} \draw[->] (\n.north|-0,-2.05) -- (\n.north);
\end{tikzpicture}
\caption{Playing, recording, checking and scoring a run in \bench. (1)~At every \benchDecideEvery{}th close, the harness builds the prompt from data published before that close. (2)~The LLM agent replies once with its buys and sells. (3)~Orders fill at the next day's open, with costs. (4)~The recorded run is checked for look-ahead and scored without calling the model again.}
\label{fig:overview}
\end{figure*}
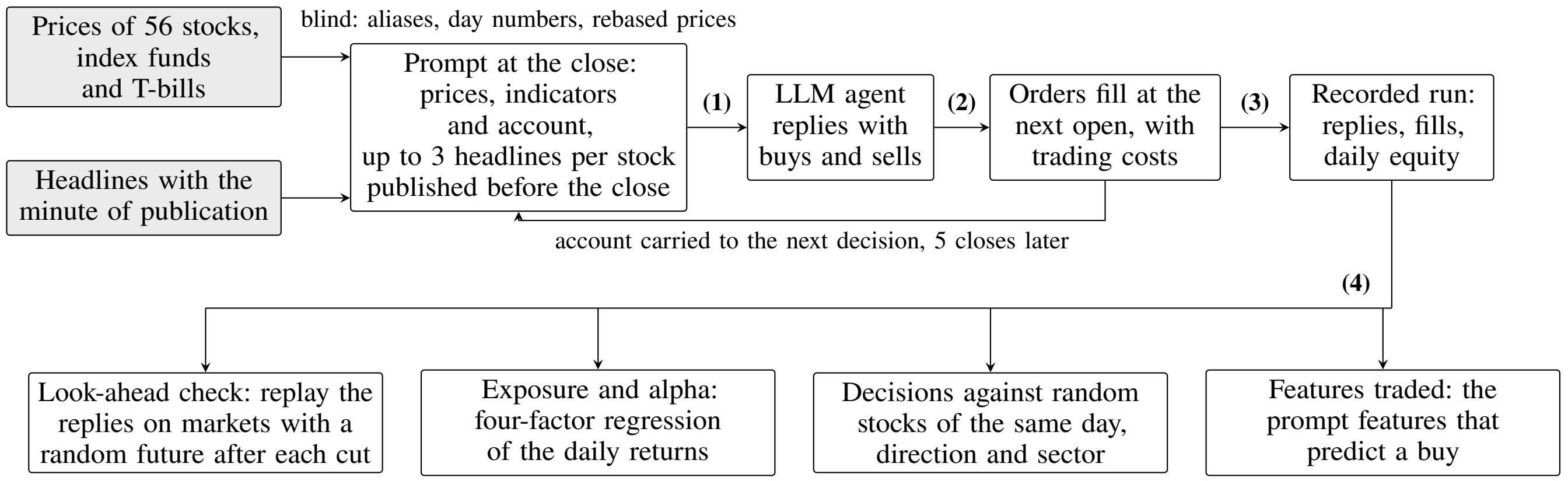

\bench{} replays three stretches of the US stock market and lets an agent trade each of them, with every headline arriving at the minute it was published. Every agent trades the same stocks on the same days under the same rules, and every run leaves a record from which anyone can recompute its scores without calling the model again. This section describes the markets, the news, a single decision, the two ways each period is played, the check that keeps the future out of the prompt, and the scores. \Cref{fig:overview} shows how a run is played, recorded, checked and scored.

\subsection{Markets and News}
\label{sec:markets}

The three periods were chosen to look different from one another (\cref{tab:periods}). The 2026 period, from \firstDayTwentySix{} to \lastDayTwentySix, is a rising market with a sell-off in March. The 2022 period is the rate-hike bear market, in which the Nasdaq 100 fell by about a third. The 2020 period holds the COVID crash, in which SPY lost a third of its value in 23 trading days, and the rebound to a new high by August. An agent that only does well when everything rises does poorly in 2022 and in the crash.

Each period ships as a set of files that never change. They hold daily total-return prices for \benchStocks~large US stocks in \benchSectors~sectors and for the index, sector and bond exchange-traded funds that the market readings use. Two years of earlier prices are included, so that slow indicators such as the 200-day average are ready on the first day. A new period is a new set of files, so a published score keeps pointing at the same data, and every record stores the SHA-256 hashes of the files it used. Uninvested cash earns the daily return of a T-bill fund, as it would in a brokerage account.

\begin{table}[t]
\centering
\caption{The three periods. SPY and the equal-weight portfolio of the benchmark's \benchStocks~stocks are held for the whole window. The equal-weight portfolio beats SPY in 2020 by a wide margin because the stock list is today's large caps.}
\label{tab:periods}
\small
\setlength{\tabcolsep}{4pt}
\begin{tabular}{@{}lrrrr@{}}
\toprule
 & & \multicolumn{2}{c}{SPY (\%)} & Equal \\
\cmidrule(lr){3-4}
Period & Days & Return & Drawdown & weight (\%) \\
\midrule
2026 rally & \daysTwentySix & \retSpyTwentySix & \ddSpyTwentySix & \retEwTwentySix \\
2022 bear market & \daysTwentyTwo & \retSpyTwentyTwo & \ddSpyTwentyTwo & \retEwTwentyTwo \\
2020 crash, rebound & \daysTwenty & \retSpyTwenty & \ddSpyTwenty & \retEwTwenty \\
\bottomrule
\end{tabular}
\end{table}

\rih{News Replayed by Publication Time.}\label{sec:replay} The news is a feed of financial headlines that name the benchmark's companies, each stored with the time it was published. In the 2026 period it also holds the company filings (Form 8-K) that the Securities and Exchange Commission (SEC) received, each with the time the SEC accepted it. The headlines come from a licensed feed and are not redistributed. A decision is taken after the close of a trading day, at 16:00 New York time or at 13:00 on the days the market closes early. An item counts for that decision only if it was published strictly before that moment. An item stamped at the close or later counts from the next trading day. Two rules make every doubtful timestamp err toward showing the news late rather than early. An item with a date and no time counts at the end of that day. A time without a zone counts as New York time, so a time that was really in UTC shows the item a few hours late, never early.

At each decision, the prompt shows at most \benchMaxNews~headlines per stock from the last 72 hours. The package groups headlines about the same event into one story and marks a story as confirmed when an SEC filing, the company's own release or two independent outlets carry it. It scores each story from $-1$ to $1$ with a finance word list, and the stock's news score weights confirmed and recent stories more. The prompt lists three stories, confirmed stories first and then the most recent, each with its confirmation status and its age in hours at the close. It also gives the stock's overall news score and any red flags, such as a fraud probe or a guidance cut. Live trading uses the same code, so a backtest and a live session treat a story the same way.

\subsection{Decisions, Blind Runs and the Training Cutoff}
\label{sec:decision}

The agent decides after every \benchDecideEvery{}th close, a weekly rhythm that suits swing trading on daily bars. Its prompt holds three parts. The market part gives the regime (risk-on, neutral, uncertain or risk-off) and the readings behind it, such as SPY's trend, the Cboe Volatility Index (VIX) and its term structure, and sector breadth. The portfolio part gives the cash share, the return since the start and every position with its weight, gain and days held. The stock part lists each of the \benchStocks~stocks with its sector, its returns over 1, 5 and 21 days, and its 2- and 14-day relative strength index (RSI). It adds the stock's distance from its 20-, 50- and 200-day averages and from its 52-week high, its average true range, its relative volume and its news.

The agent answers with a list of orders in JSON: buy a stock up to a share of equity, sell a fraction of a position, or move a position to a target weight. An empty list holds everything. The portfolio is long only, holds at most \benchMaxPositions~stocks, starts with \$\benchCash, and fills every order at the next trading day's open. Each fill pays a realistic cost of 5~bp plus a tenth of the overnight gap between the close and the open. A market-impact term adds the stock's daily volatility times the square root of the order's share of its average daily volume, and no order may exceed 1\% of that volume. A reply that cannot be parsed, or a model call that fails, is recorded and the agent holds its portfolio for that step, and an order for a stock that does not exist is recorded and skipped.

\rih{Blind and Named Runs.}\label{sec:blind} Each period is played twice. A named run shows everything. A blind run replaces tickers with aliases (S01, S02, and so on), dates with day numbers, and prices with prices rebased to 100 on the first day. Inside the headlines, company names and tickers become the same aliases, and the companies' best-known people and products are masked. Dates and years are removed, and per-share dollar amounts are rebased like the prices, so that a price target keeps its distance from the price without giving the price level away. A seed fixes the aliases, so the same seed gives the same blind market.

Blind runs lower the chance that a model recognizes a stock or a date. The gap between a named and a blind run of the same seed shows how much the identities change what the model does. Each model also carries its training cutoff, taken from the vendor's model page. A period that starts after the cutoff is clean. If the cutoff falls inside a period, the run is also scored from the last close on or before the cutoff. Scores from periods that the model may have read about are listed apart from the clean scores.

\subsection{Checking and Scoring a Run}
\label{sec:check}

\rih{The Look-Ahead Check.} The design keeps the future out of reach: indicators and price history reach the agent as copies that end at the decision day, and news is filtered by publication time. The look-ahead check tests the result instead of trusting the design. It replays the model's recorded replies on three copies of the market in which every price and every headline after a cut day is replaced by a random one. Every observation up to the cut must stay identical, byte for byte. A pass shows that nothing after the cut reached the prompt, without calling the model again.

The check only looks at the observations up to each cut, so where the cuts fall matters. A leak that lasts one evening, such as a calendar-date join, changes the observation of a decision day only when the cut falls on that day. With a decision every fifth close and cuts at fixed fractions of the period, the cuts may miss every decision day. We test this in \cref{sec:checkresults} by planting four kinds of leak and running the check with the package's cuts and with the cuts moved to decision days.

\label{sec:scoring}A year's return is one draw from one market, so the benchmark scores each run in three more ways.

\rih{Exposure and Selection.} The run's daily return over T-bills is regressed on four factors built from the frozen prices~\cite{fama1993common,carhart1997persistence,jegadeesh1993returns}. They are the market (SPY minus T-bills), size (small caps minus SPY), growth (the Nasdaq 100 minus SPY) and momentum, which is the best third minus the worst third of the stock list by the return from six months to one month back. Standard errors are Newey-West with four lags~\cite{newey1987simple}. The part that the exposures explain is the return the agent would have earned by holding the factors, and the intercept, alpha, is what is left for selection.

\rih{Decisions.} Every buy and every sell is scored by the stock's return minus SPY's over the next 1, 5 and 20 trading days, from the open of the fill day. A buy is credited with the excess return it earned and a sell with the excess return it avoided. The $t$-statistics treat decisions on the same day as one cluster and weight overlapping windows by the days they share. The rank information coefficient (IC) asks whether the agent's actions on a decision day ranked the stocks by their next returns.

\rih{Null Distributions.} A decision statistic means little without knowing what a random choice would have earned on the same days. For every decision we swap the stock for a random other stock of the same sector, keep the day and the direction, and repeat 1,000 times. For every run we also keep its daily stock weights and swap each holding for a random stock of the same sector, 200 times. Both give the distribution of the agent's statistics if it picked stocks at random within its chosen sectors.

\rih{Entries That Replay.} A leaderboard entry is one JSON file that holds the run's settings, the hashes of its data and records, every reply of the model and the scores. Replaying the replies through the backtest must give the same return, Sharpe ratio, drawdown and turnover, so a submitted entry cannot claim a result that its own orders do not produce. We replayed all \replayRuns~runs of the main study and recovered every recorded fill.

\section{Experimental Setup}
\label{sec:setup}

We test six models from four vendors (\cref{tab:models}). They range from a 3-billion-parameter open-weight model to a mid-size commercial model. Their training cutoffs fall before, inside or after the 2026 period, so the same benchmark has clean and possibly remembered scores for each of them. Every model is called with the package's system prompt (Appendix~\ref{app:prompt}). The prompt describes the task and the reply format and gives no trading style, so differences between runs come from the models and the market.

\begin{table}[t]
\centering
\caption{The six models and their training cutoffs as stated by their vendors. The 2026 period is clean when it starts after the cutoff and partly clean when the cutoff falls inside it. Every cutoff falls after 2022, so every model may remember the 2022 and 2020 periods.}
\label{tab:models}
\small
\begin{tabular}{@{}lllc@{}}
\toprule
Model & Vendor & Cutoff & 2026 \\
\midrule
GPT-5.6 Luna & OpenAI & Feb 2026 & partly \\
Gemini 3.5 Flash & Google & Jan 2025 & clean \\
Claude Haiku 4.5 & Anthropic & Jul 2025 & clean \\
Claude Sonnet 4.6 & Anthropic & Jan 2026 & partly \\
Claude Haiku 5.5 & Anthropic & Jun 2026 & partly \\
Llama 3.2 3B & Meta & Dec 2023 & clean \\
\bottomrule
\end{tabular}
\end{table}

\rih{Runs.} A run is one model trading one period, blind or named, with one seed. The seed fixes the blind aliases. The analyses below use the \nRuns~runs of the six models that had finished when we froze the analysis (Appendix~\ref{app:runs} lists them). Most of them come from the benchmark's leaderboard. There, GPT-5.6 Luna, Gemini 3.5 Flash and Llama 3.2 3B have all three periods, blind and named, with seeds 0, 1 and 2, and the three Claude models have seed 0. The leaderboard runs were made over several weeks, with small changes to the code in between. We therefore ran GPT-5.6 Luna again with all three seeds, and Gemini 3.5 Flash with seed 0, on one frozen version of the code. These reruns count as runs of the main study. The leaderboard's Gemini 3.5 Flash runs with seeds 1 and 2 were made on the same frozen version. Of the \nSteps~steps in the main study, \noReplySteps~had no usable reply, because the model call failed or the reply held no valid JSON. All of them were in runs of Llama 3.2 3B, one in each of \noReplyRuns~runs, and the agent held its portfolio at those steps.

\rih{Ablations.} Three sets of runs change one thing in the blind runs of GPT-5.6 Luna and Gemini 3.5 Flash on the frozen code, each with three seeds: no news at all, another stock's news, and news joined by calendar date. Each ablation run is paired with its baseline run, the blind run on the frozen code with the same model, period and seed (\cref{sec:news}). Two more kinds of experiment come from the benchmark's own studies. Its prompt studies change one thing at a time in the system prompt, and its short-selling study lets accounts sell short and borrow on margin (\cref{sec:robustness}).

\rih{References.} Every period also runs five textbook swing-trading rules under the same costs and limits. They are trend following on moving-average crossovers, RSI(2) mean reversion, breakouts above the 20-day high, buying pullbacks to the 50-day average, and buying after positive headlines. Next to them sit SPY and the equal-weight portfolio of the \benchStocks~stocks. All statistics come from the frozen market alone.

\section{Sources of the Agents' Returns}
\label{sec:returns}

A return alone cannot say whether the agents earned it by picking stocks or by holding the market and leaning toward some styles. This section first compares the returns with simple references and splits each run's return into factor exposure and alpha. It then scores every decision against random choices and identifies the features of the stocks that the agents bought.

\subsection{Returns, Exposure and Alpha}
\label{sec:leaderboard}

In their blind runs the agents trailed SPY in the rising market and lost less than SPY in the falling one (\cref{tab:returns}). In 2026 they made between \retLowTwentySix\% and \retHighTwentySix\% on average against \retSpyTwentySix\% for SPY, and in the 2020 rebound only Sonnet 4.6 beat SPY. In 2022 all \nModels~models lost less than SPY on average, but they did so the way the rules did, by holding less stock. Their market betas were well below one, between \betaMktLo{} and \betaMktHi{} across all runs, so in a market that returned \retSpyTwentyTwo\%, a portfolio that was half in T-bills lost about half as much.

\begin{figure*}[t]
\centering
\includegraphics[width=\textwidth]{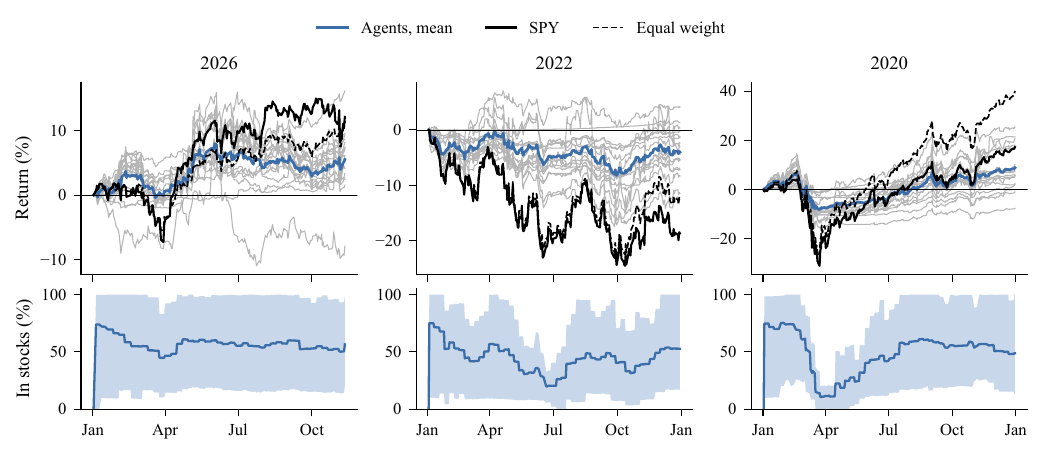}
\caption{Blind runs over the three periods. Top: each run's return (gray), the mean over runs (blue), SPY, and the equal-weight portfolio of the \benchStocks~stocks. Bottom: the share of equity held in stocks, as the mean over runs (line) and the range between the 10th and the 90th percentile of runs (band).}
\label{fig:equity}
\end{figure*}

The agents' betas were low because they kept much of their money in cash (\cref{fig:equity}). On an average day, they held \expoTwentySix\% of their equity in stocks in 2026, \expoTwentyTwo\% in 2022 and \expoTwenty\% in 2020, and single runs ranged from \expoMinAll\% to \expoMaxAll\%. In 2020 they also moved their cash at the wrong times. At the market's peak on 19 February, the agents held \expoPeakTwenty\% of their equity in stocks on average, and at its low on 23 March they held \expoLowTwenty\%. At the end of June, after most of the rebound, they held \expoJuneTwenty\%. They sold most of their stocks during the crash and bought them back only after much of the rebound.

\begin{finding}[label=fnd:crash]
In the 2020 crash, the agents sold near the bottom and bought back late. On average they had \expoPeakTwenty\% of their equity in stocks at the market's peak on 19 February, \expoLowTwenty\% at the low on 23 March, and only \expoJuneTwenty\% at the end of June, after most of the rebound.
\end{finding}

The rule-based references make the same point from the other side. Trend following, a rule with no forecast at all, made \retTrendTwentySix\% in 2026 and \retTrendTwenty\% in 2020, more than any agent, because it held winners for months. The equal-weight portfolio of today's large caps made \retEwTwenty\% in 2020, more than every agent and every rule but trend following, which shows how much of any 2020 result comes from the stock list itself.

\begin{finding}[label=fnd:trend]
A simple trend-following rule made more money than every LLM agent. It made \retTrendTwentySix\% in 2026 and \retTrendTwenty\% in 2020, against \retSpyTwentySix\% and \retSpyTwenty\% for SPY, while no model averaged more than \retHighTwentySix\% in 2026.
\end{finding}

\begin{table*}[t]
\centering
\caption{Blind runs of each model: number of runs (leaderboard runs and frozen-code reruns together), and their mean total return over the period (\%), market beta and four-factor alpha (\% a year). No run's alpha is significant (largest $|t|$ \alphaMaxAbsT). The five swing-trading rules and the buy-and-hold references trade the same \benchStocks~stocks under the same costs; for them only the return is shown.}
\label{tab:returns}
\small
\setlength{\tabcolsep}{3.5pt}
\begin{tabular}{@{}lrrrrrrrrrrrr@{}}
\toprule
 & \multicolumn{4}{c}{2026 rally} & \multicolumn{4}{c}{2022 bear market} & \multicolumn{4}{c}{2020 crash and rebound} \\
\cmidrule(lr){2-5}\cmidrule(lr){6-9}\cmidrule(lr){10-13}
 & Runs & Return & Beta & Alpha & Runs & Return & Beta & Alpha & Runs & Return & Beta & Alpha \\
\midrule
GPT-5.6 Luna & 6 & +7.8 & 0.35 & +3.5 & 6 & $-$3.7 & 0.29 & $-$3.0 & 6 & +0.9 & 0.30 & $-$8.4 \\
Gemini 3.5 Flash & 4 & +6.3 & 0.70 & $-$1.6 & 4 & $-$8.4 & 0.46 & $-$8.9 & 4 & +13.7 & 0.47 & $-$9.2 \\
Haiku 4.5 & 1 & +7.2 & 0.59 & $-$1.2 & 1 & $-$5.3 & 0.58 & $-$1.0 & 1 & +16.9 & 0.50 & $-$2.3 \\
Sonnet 4.6 & 1 & +8.1 & 0.59 & +1.8 & 1 & +0.4 & 0.55 & $-$0.5 & 1 & +25.4 & 0.34 & +3.6 \\
Haiku 5.5 & 1 & +2.4 & 0.23 & $-$3.2 & 1 & +1.4 & 0.01 & $-$0.0 & 1 & +7.4 & 0.23 & $-$2.6 \\
Llama 3.2 3B & 3 & $-$0.6 & 0.19 & $-$8.7 & 3 & $-$1.6 & 0.25 & +2.7 & 3 & +10.6 & 0.19 & +7.7 \\
\midrule
Trend following &  & +19.0 &  &  &  & $-$5.9 &  &  &  & +67.4 &  &  \\
RSI(2) reversion &  & +4.9 &  &  &  & $-$8.6 &  &  &  & +6.3 &  &  \\
20-day breakout &  & $-$6.5 &  &  &  & $-$15.1 &  &  &  & +27.6 &  &  \\
50-day pullback &  & +1.1 &  &  &  & $-$4.2 &  &  &  & +2.2 &  &  \\
Positive headlines &  & +2.0 &  &  &  & $-$16.3 &  &  &  & +23.4 &  &  \\
SPY &  & +12.1 &  &  &  & $-$18.6 &  &  &  & +17.2 &  &  \\
Equal weight &  & +11.4 &  &  &  & $-$12.4 &  &  &  & +40.1 &  &  \\
\bottomrule
\end{tabular}
\end{table*}

\begin{figure*}[t]
\centering
\includegraphics[width=\textwidth]{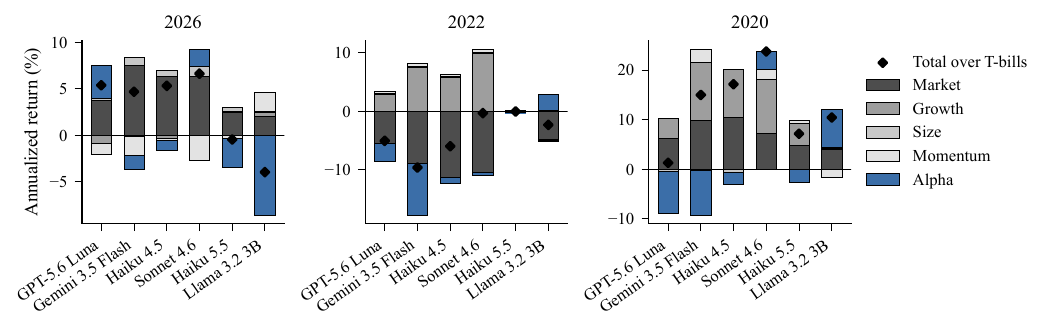}
\caption{Each model's annualized return over T-bills in its blind runs (diamonds), split into the parts that its exposures to the market, growth, size and momentum factors explain and the alpha that remains. Market and growth exposure make up most of every bar, and the alpha changes sign from period to period.}
\label{fig:attribution}
\end{figure*}

\label{sec:exposure}The factor regression explains most of what the agents earned (\cref{fig:attribution}). Across the \nRuns~runs, blind and named, not one alpha was significant, and the largest $|t|$ was \alphaMaxAbsT. A single run has about a year of daily returns, so its alpha has a standard error of about \alphaSeMedianPct\% a year (median over runs), and only an alpha of roughly twice that would be significant. The mean over all runs is the stronger test. The alphas averaged \alphaMeanPct\% a year, with a 95\% interval from \alphaLoPct\% to \alphaHiPct\% when we resample the \alphaCells~combinations of model and period, since seeds and reruns of the same model and period are not independent. The interval barely reaches above zero, so after their exposures are accounted for, the agents' stock choices added nothing and may have cost them money. In the median run, the exposures account for \exposureShareMedian\% of the absolute size of the return's parts, and the median $R^2$ is \rTwoMedian.

\begin{finding}[label=fnd:alpha]
None of the agents beat the market once we took out what the market and style factors paid. Across \nRuns~runs in three markets, no run had a significant four-factor alpha (the largest $|t|$ was \alphaMaxAbsT), and the average alpha was \alphaMeanPct\% a year.
\end{finding}

The exposures differ in a way that follows each market. In 2026 most agents leaned away from growth stocks and earned little from it. In 2022 that same lean against growth saved them money as the Nasdaq 100 fell, and in 2020 the agents that leaned into growth, such as Sonnet 4.6, made the most. An exposure is a bet on a factor, and a bet that pays in one year costs in another.

The reruns on one frozen version of the code give the same result, and none of their \rerunRuns~alphas was significant either (largest $|t|$ \rerunMaxAbsT). They also show how much a single run varies. Over \rerunPairs~pairs, a rerun's total return differed from that of the leaderboard run with the same model, period and seed by \rerunAbsDiff~percentage points on average. Two seeds of the same model in the same period differed by \seedAbsDiff~points, over \seedPairs~pairs. In 2026, the average returns of the best and the worst model were \modelSpreadTwentySix~points apart, so the return of a single run cannot rank most of these models.

\begin{finding}[label=fnd:seeds]
Running a model once is not enough to rank it. Two runs of the same model that differed only in the random seed ended \seedAbsDiff~points apart on average, more than half of the \modelSpreadTwentySix-point gap between the best and the worst model of 2026.
\end{finding}

\subsection{Decisions and the Features Behind Them}
\label{sec:decisions}

\begin{figure}[t]
\centering
\includegraphics[width=\columnwidth]{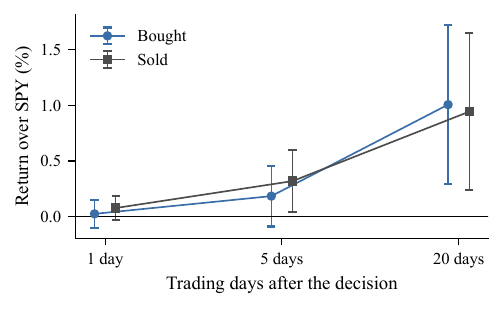}
\caption{Return of the traded stock minus SPY's after the agents' buys and sells, from the open of the fill day, with 95\% intervals that treat decisions on the same day as one cluster. After 20 trading days, the stocks the agents sold had done as well as the stocks they bought.}
\label{fig:horizons}
\end{figure}

A year of returns is one draw, but a run makes hundreds of decisions, each of which can be checked. Over the \nDecisions~buys and sells in our runs, the stock's return minus SPY's over the next five trading days averaged \decMeanPct\%, with a 95\% interval from \decLoPct\% to \decHiPct\%. Of these decisions, \decHit\% came out ahead. The buys earned \decBuyMeanPct\% ($t = \decBuyT$). The sells did worse than nothing. The stocks that the agents sold went on to beat SPY by \decSellAbsPct\% on average ($t = \decSellStockT$), so the agents tended to sell stocks just before they recovered. The same holds at longer horizons (\cref{fig:horizons}). Twenty trading days after the decision, the stocks that the agents bought had beaten SPY by \horBuyTwenty\% ($t = \horBuyTwentyT$), and the stocks that they sold by \horSellTwenty\% ($t = \horSellTwentyT$). Both groups beat SPY because the stock list holds today's large caps, and the choice between the stocks to buy and the stocks to sell added nothing.

\begin{figure*}[t]
\centering
\includegraphics[width=\textwidth]{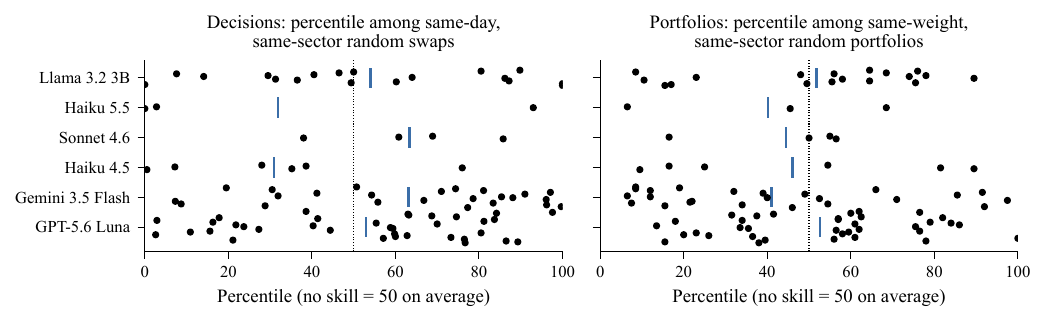}
\caption{Each run among random choices. Left: the run's mean five-day excess return per decision, as a percentile of 1,000 versions with each stock swapped for a random stock of the same sector on the same day. Right: the run's return as a percentile of 200 portfolios with the same daily weights and each holding swapped for a random stock of the same sector. Dots are runs, bars are means; no skill gives 50 on average.}
\label{fig:nulls}
\end{figure*}

To know what a random choice would have earned on the same days, we swapped each decision's stock for a random stock of the same sector, keeping the day and the direction (\cref{fig:nulls}). Without skill, a run's mean decision would have a percentile rank of 50 among these swaps on average. The agents' runs had a mean percentile rank of \nullDecPct. Of the \nullRuns~runs, \nullDecTop~were above the 95th percentile and \nullDecBottom~below the 5th, a wider spread than chance gives in both directions and no lean toward the top. The same holds for whole portfolios. With each holding swapped for a random stock of the same sector and the weights kept, the agents' returns had a mean percentile rank of \nullPortSectorPct, and with stocks from any sector, of \nullPortAnyPct. Choosing sectors added a little, and choosing stocks within sectors added nothing.

\begin{finding}[label=fnd:decisions]
The agents' trades were no better than picking a random stock from the same sector on the same day. The stocks they sold even went on to beat SPY by \decSellAbsPct\% over the next five days ($t = \decSellStockT$), so they tended to sell too early.
\end{finding}

\begin{figure*}[t]
\centering
\includegraphics[width=\textwidth]{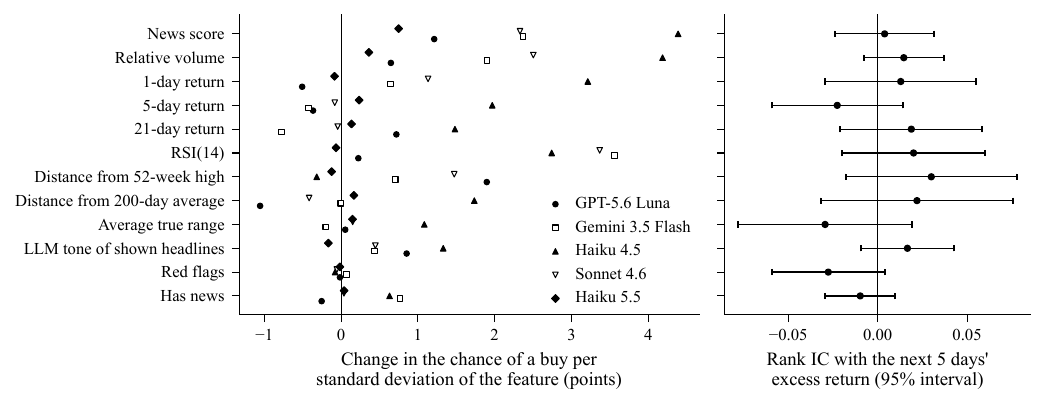}
\caption{Left: how each feature in the prompt changes the chance that a model buys a stock it does not hold, per standard deviation of the feature, in blind runs. All \nModelsFit~models that traded actively buy stocks with a high news score. Llama 3.2 3B, which made too few decisions to fit, is left out. Right: each feature's rank correlation with the next five days' return over SPY, measured from the next open on the same decision days. None is significant.}
\label{fig:factors}
\end{figure*}

\label{sec:policy}The agents did not pick winners, so we asked what they picked instead. We replayed every run to recover exactly what the model was shown at each step, and for every step and every stock it did not hold we asked whether the model bought it. We regressed that choice on the features in the prompt, each scaled across the stocks of the step. The linear regression gives the change in the chance of a buy per standard deviation of a feature (\cref{fig:factors}, left), with standard errors clustered by decision day.

One feature drives buying for all \nModelsFit~models that traded more than a few times: the news score. Llama 3.2 3B is left out, because it made only \llamaDecMin~to \llamaDecMax~decisions per run and usually bought one stock and held it. In all \newsScoreCells~combinations of model and mode (the \nModelsFit~models, blind and named, with Claude Haiku 5.5 run blind only), its effect is positive and significant ($t$ from \newsScoreTLo{} to \newsScoreTHi). One standard deviation of news score raises the chance of a buy by \newsScoreCoefLo{} to \newsScoreCoefHi~points, against a base rate of \newsScoreBase\% per stock and step. Relative volume, the other sign of attention, raises it significantly in \relVolPosSig~of \newsScoreCells~combinations. On the sell side, the agents sell the stocks that are furthest below their 52-week highs, with a negative effect in \highNeg~of \highCells~combinations.

These features do not pay (\cref{fig:factors}, right). On the decision days of the runs, the news score had a rank correlation of \icNews{} ($t = \icNewsT$) with the next five days' excess return, and relative volume had \icVol{} ($t = \icVolT$). The tone that a language model gave the headlines in the prompt had \icLlm{} ($t = \icLlmT$). No feature in the prompt reached $|t| = 2$. The agents trade attention, and among the largest US stocks attention had no measurable relation to the next week's return. Adding news factors to the return regression changes little. We built three long-short factors from the news as it was known at each close: the top third minus the bottom third of the stocks by news score, by the tone that GPT-5.6 Luna gave the day's headlines, and by the number of headlines against its 20-day average. Over \nfRuns~runs, the news-score factor had a significant positive loading in \nfScorePos~runs and a significant negative loading in \nfScoreNeg, and the tone factor had \nfTonePos~positive and \nfToneNeg~negative ones. The median $R^2$ was \nfRTwoSeven{} with the news factors and \nfRTwoFour{} without them, and no alpha became significant (largest $|t|$ \nfAlphaMaxAbsT). The agents' taste for news is real, but it is too small a part of their portfolios to show up in their returns.

\begin{finding}[label=fnd:buynews]
Every model that traded actively bought the stocks with the best news, but good news did not predict the next week's return. A higher news score made a buy more likely for all \nModelsFit~of these models ($t$ from \newsScoreTLo{} to \newsScoreTHi), yet its rank correlation with the next five days' return over SPY was \icNews.
\end{finding}

\section{The Value of the News}
\label{sec:news}

All \nModelsFit~models that traded more than a few times bought the stocks with the most positive news (\cref{sec:policy}). If the news says something about the next week's returns, the agents should therefore earn more with it than without it. This section tests the news in five ways. Three ablations change only the news in the agent's prompt and run on the benchmark's frozen code. In them, GPT-5.6 Luna and Gemini 3.5 Flash, each with three seeds, traded the three periods blind with no headlines, with another stock's headlines, and with headlines joined by calendar date. Each ablation run is compared with the blind run of the same model, period and seed that received the headlines by publication time. We then ask whether the headlines lowered the agents' risk, which kinds of headlines moved prices once they were public, how fast prices took them in, and whether the benchmark's look-ahead check catches a calendar-date join.

\subsection{Three Ablations of the News}
\label{sec:nonews}

\rih{No News and Another Stock's News.} The no-news runs give the agent the same prices, indicators and account as the baseline, with no news block in any stock's line. The swapped runs keep the news blocks but move them between stocks. On each decision day, every stock that has news receives the block of another such stock, chosen by a seeded random permutation in which no stock keeps its own. The headlines then name the receiving stock's alias. The agent therefore receives news of the usual kind and amount, about the wrong company. If the agents earn anything by reading the news, both ablations should lower their returns and worsen their decisions.

\begin{figure*}[t]
\centering
\includegraphics[width=\textwidth]{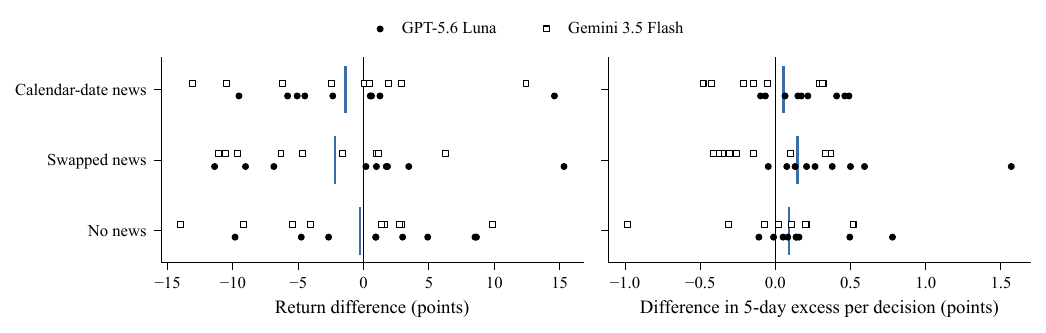}
\caption{Each ablation run minus the baseline run of the same model, period and seed. Left: difference in total return. Right: difference in the mean five-day return over SPY per decision. Filled markers: GPT-5.6 Luna. Open markers: Gemini 3.5 Flash. Bars: means over pairs.}
\label{fig:ablations}
\end{figure*}

Neither ablation made the agents worse (\cref{tab:ablation} and \cref{fig:ablations}). Without headlines, the total return differed from the paired baseline run's by \abNoNewsRet~percentage points on average (95\% interval \abNoNewsRetLo{} to \abNoNewsRetHi), and the mean five-day return per decision by \abNoNewsDec~points (\abNoNewsDecLo{} to \abNoNewsDecHi). With another stock's headlines, the two differences were \abShufRet~points (\abShufRetLo{} to \abShufRetHi) and \abShufDec~points (\abShufDecLo{} to \abShufDecHi). All four intervals include zero. The agents did use the news, because \abBaseNewsShare\% of the baseline runs' decisions gave news as their reason. However, what the headlines said about a company did not make the agents' decisions better than those made without news or with the wrong company's news.

\begin{table}[t]
\caption{Ablation runs minus the baseline run of the same model, period and seed: the mean difference over pairs, with a 95\% bootstrap interval. Return: total return over the period, in percentage points. Decision: mean five-day return over SPY per decision, in percentage points.}
\label{tab:ablation}
\centering
\footnotesize
\begin{tabular}{@{}lrcc@{}}
\toprule
Ablation & Pairs & Return & Decision \\
\midrule
No news & 18 & $-$0.3 [$-$3.4, +2.7] & +0.09 [$-$0.08, +0.24] \\
Swapped news & 18 & $-$2.2 [$-$5.3, +1.1] & +0.15 [$-$0.05, +0.38] \\
Calendar date & 18 & $-$1.4 [$-$4.4, +1.9] & +0.05 [$-$0.08, +0.19] \\
\bottomrule
\end{tabular}
\end{table}

\rih{News Joined by Calendar Date.}\label{sec:calendar} A backtest that joins news to prices by calendar date shows an agent that decides at the close every headline dated that day, including those published in the evening after the close. The calendar-date runs make this one change: a headline counts for a decision day if it was published before midnight New York time rather than before the 4~p.m. close. Orders still fill at the next open, as in every other run, so the evening's news is already in the price the agent pays. A backtest that also trades from close to close would instead credit the agent with the move from the decision day's close. To separate the two, we score every decision of the baseline and calendar-date runs twice, from the next open and from the decision day's close.

The extra evening headlines did not help the agents either (\cref{tab:twobytwo}). Measured from the next open, the decisions of the calendar-date runs earned \ttCalAllOpen\% over five days ($t = \ttCalAllOpenT$), close to the \ttPubAllOpen\% of the baseline runs. A calendar-date join can only change the decisions in stocks with a headline published after the close. These decisions earned \ttCalAfterOpen\% from the next open ($t = \ttCalAfterOpenT$) and \ttCalAfterClose\% from the close ($t = \ttCalAfterCloseT$), so even a close-to-close backtest would not have credited the agents with a significant edge. Their total return differed from the paired baseline runs' by \abCalRet~points (\abCalRetLo{} to \abCalRetHi), and the share of their decisions that gave news as the reason barely changed (\abCalNewsShare~percentage points).

\begin{table}[t]
\caption{Mean five-day return over SPY per decision (\%), with its $t$-statistic in parentheses, for the baseline runs (news by publication time) and the calendar-date runs. Open: measured from the next open, where the benchmark fills orders. Close: measured from the decision day's close, as a close-to-close backtest would. After-close news: decisions in stocks with a headline published between that day's close and midnight.}
\label{tab:twobytwo}
\centering
\footnotesize
\begin{tabular}{@{}lcccc@{}}
\toprule
 & \multicolumn{2}{c}{All decisions} & \multicolumn{2}{c}{After-close news} \\
\cmidrule(lr){2-3}\cmidrule(lr){4-5}
News & Open & Close & Open & Close \\
\midrule
Publication time & $-$0.04 & $-$0.00 & $-$0.15 & +0.06 \\
 & ($-$0.4) & (0.0) & ($-$0.4) & (+0.2) \\
Calendar date & +0.02 & +0.08 & $-$0.00 & +0.53 \\
 & (+0.2) & (+0.6) & (0.0) & (+1.5) \\
\bottomrule
\end{tabular}
\end{table}

\subsection{Risk With and Without the News}
\label{sec:risk}

What the headlines said did not lower the agents' risk. News could still help an agent that earns no alpha, if it kept the agent out of a stock before a fall or out of the market before a crash, so we compared the risk of each ablation run with that of its paired baseline run, using the run's daily equity. The measures are the average share of equity held in stocks, the worst loss over five trading days, and the largest fall from a peak, which finance calls the maximum drawdown. We also computed the annualized volatility of daily returns, the same measure over losing days only, the mean of the worst 5\% of daily returns, known as the conditional value at risk (CVaR), and the market beta. \Cref{tab:risk} gives the mean difference over the \riskNoNewsPairs~pairs of each ablation, with a 95\% bootstrap interval.

Without headlines, the agents lost more in their worst week, but what the headlines said did not matter. Their worst five-day loss was \riskNoNewsWorst~points larger than in the paired baseline runs (95\% interval \riskNoNewsWorstLo{} to \riskNoNewsWorstHi), and it was larger in \riskNoNewsWorstPos~of the \riskNoNewsPairs~pairs. The drawdown, both volatilities, the CVaR and the beta were also higher on average, but none of these intervals excludes zero. Their share of equity in stocks did not change measurably (\riskNoNewsExpo~points, \riskNoNewsExpoLo{} to \riskNoNewsExpoHi), because the two models moved in opposite directions. Without headlines, GPT-5.6 Luna held \riskNoNewsExpoLunaAbs~points more of its equity in stocks, and Gemini 3.5 Flash held \riskNoNewsExpoGeminiAbs~points less. With another stock's headlines, no risk measure changed (\cref{tab:risk}), so what the headlines said about the company did not make the agents safer.

\begin{table}[t]
\caption{Risk of each ablation run minus that of the baseline run of the same model, period and seed: the mean over \riskNoNewsPairs~pairs, in percentage points (beta in its own units), with a 95\% bootstrap interval below. Share in stocks: average share of equity held in stocks. Volatility: annualized, of daily returns; downside volatility counts losing days only. CVaR: mean of the worst 5\% of daily returns.}
\label{tab:risk}
\centering
\footnotesize
\setlength{\tabcolsep}{4pt}
\begin{tabular}{@{}lccc@{}}
\toprule
Measure & No news & Swapped news & Calendar date \\
\midrule
Share in stocks & +3.8 & $-$1.8 & +0.7 \\
 & [$-$2.2, +9.6] & [$-$5.2, +1.6] & [$-$2.7, +4.0] \\
Worst five-day loss & +0.8 & +0.6 & +1.1 \\
 & [+0.2, +1.5] & [$-$0.4, +1.4] & [+0.4, +1.8] \\
Maximum drawdown & +1.1 & +1.8 & +2.0 \\
 & [$-$0.4, +2.8] & [$-$0.2, +3.7] & [+0.4, +3.4] \\
Volatility & +0.7 & +0.3 & +1.1 \\
 & [$-$0.6, +1.9] & [$-$1.0, +1.4] & [+0.0, +2.1] \\
Downside volatility & +0.4 & +0.3 & +0.9 \\
 & [$-$0.6, +1.4] & [$-$0.9, +1.1] & [+0.1, +1.8] \\
Daily 5\% CVaR & +0.08 & +0.05 & +0.21 \\
 & [$-$0.13, +0.29] & [$-$0.22, +0.26] & [+0.03, +0.38] \\
Market beta & +0.02 & $-$0.00 & +0.03 \\
 & [$-$0.03, +0.07] & [$-$0.05, +0.04] & [$-$0.01, +0.08] \\
\bottomrule
\end{tabular}
\end{table}

Headlines seen early added risk. The calendar-date runs, which saw the evening's headlines before deciding, carried more risk than the baseline runs on five of the seven measures, with intervals above zero. Their worst five-day loss was \riskCalWorst~points larger (\riskCalWorstLo{} to \riskCalWorstHi), in \riskCalWorstPos~of the \riskCalPairs~pairs, and their maximum drawdown \riskCalMdd~points larger (\riskCalMddLo{} to \riskCalMddHi), while their share in stocks did not change (\riskCalExpo~points). The 2020 crash is the clearest test of whether the news helped the agents get out in time. From the market's peak on 19 February to its low on 23 March, the agents with headlines lost \riskCrashWith\% and those without lost \riskCrashWithout\% (\riskCrashPairs~pairs, with a 95\% interval for the difference of \riskNoNewsCrashLo{} to \riskNoNewsCrashHi~points). The agents without headlines entered the crash with more of their equity in stocks (\riskPeakWithout\% against \riskPeakWith\%) but held less at the low (\riskLowWithout\% against \riskLowWith\%), and the two groups lost about as much.

Bad news about a stock the agents held did not make them sell it either. For every stock a run held on a decision day, we checked whether a bad new-fact headline about it had been published in the 72 hours before the decision, using the event labels of \cref{sec:events}. With headlines, the agents sold \riskSellBadNews\% of the stocks with such news, against \riskSellOtherNews\% of the other stocks they held. Without headlines, they sold \riskSellBadNoNews\% and \riskSellOtherNoNews\%, although the bad news was then hidden from them, so the higher sell rate of these stocks came from what their prices showed. Paired by run, the headlines changed the sell rate of stocks with bad news by \riskSellDiff~points (\riskSellDiffLo{} to \riskSellDiffHi). In the policy regressions of \cref{sec:policy}, a lower news score made a sale more likely in \riskSellNewsNeg~of \riskSellNewsCells~cells of model and condition, but in none with $|t| \ge 2$.

Selling these stocks would not have lowered the agents' losses. Over the next 20 trading days, the held stocks with bad news beat SPY by \riskBadTwenty\% ($t = \riskBadTwentyT$) instead of falling behind it, and those the agents sold did no worse than those they kept (\riskSoldTwenty\% against \riskKeptTwenty\%). For the largest US stocks in our periods, a bad headline was not a reason to sell.

\begin{finding}[label=fnd:badnews]
Reading a bad headline did not make the agents sell, and selling would not have helped. They sold \riskSellBadNews\% of the stocks they held after a bad headline, about the same as the \riskSellBadNoNews\% they sold when the headlines were hidden, and those stocks went on to beat SPY by \riskBadTwenty\% over the next 20 trading days ($t = \riskBadTwentyT$).
\end{finding}

How many headlines the agent sees changed its risk little. In the second prompt study of \cref{sec:prompts2}, GPT-5.6 Luna traded with no headlines, while the news score and the red flags stayed in the prompt, or with six headlines per stock instead of three (\cref{tab:prompts}). Without headlines, its share in stocks was \plNewsZeroExpoAbs~points lower (95\% interval \plNewsZeroExpoLo{} to \plNewsZeroExpoHi), and its maximum drawdown, worst five-day loss and volatility did not change measurably. With six headlines, its maximum drawdown was \plNewsSixMddAbs~points smaller (\plNewsSixMddLo{} to \plNewsSixMddHi) and its volatility \plNewsSixVolAbs~points lower (\plNewsSixVolLo{} to \plNewsSixVolHi), each in \plNewsSixMddDown~of the \plPairs~pairs. Its share in stocks and its worst five-day loss did not change measurably.

\subsection{Headlines That Move Prices}
\label{sec:events}

The news score in the prompt adds up the tone of a stock's recent headlines, and it had no relation to the next week's returns (\cref{sec:policy}). The package's own news-factor study finds the same for single days. A dictionary tone score had a rank correlation of \lexIcPub{} ($t = \lexIcPubT$) with the next day's return from the next open. The correlation rises to \lexIcCal{} ($t = \lexIcCalT$) if the same headlines are joined by calendar date and the return runs from close to close. A tone score from GPT-5.6 Luna did better, with \llmIcPub{} ($t = \llmIcPubT$) by publication time and \llmIcCal{} ($t = \llmIcCalT$) by calendar date. However, its top-ranked stocks changed so often that \llmTurnoverPct\% of them were replaced each day. A cost of \llmBreakEvenBp~basis points per side would erase the spread between the top and bottom groups.

A tone score may hide events that do matter. GPT-5.6 Luna therefore labeled each headline with one of eleven event types: earnings results, guidance, analyst ratings, mergers, legal and regulatory actions, products and contracts, executive changes, capital actions, sector stories, trading activity, and other. It also labeled the direction for the stock (good, bad or neutral) and whether the headline reports a new fact rather than commentary or a recap of price moves. The labeling model was given only the headline. An event is a stock and day with at least one new-fact headline of a given type and direction. Its return is the stock's return minus SPY's, signed by the direction. We measure it from the first open after the headline, which a trade could get, and from the close of the headline's calendar date, as a calendar-date join would.

\begin{figure*}[t]
\centering
\includegraphics[width=\textwidth]{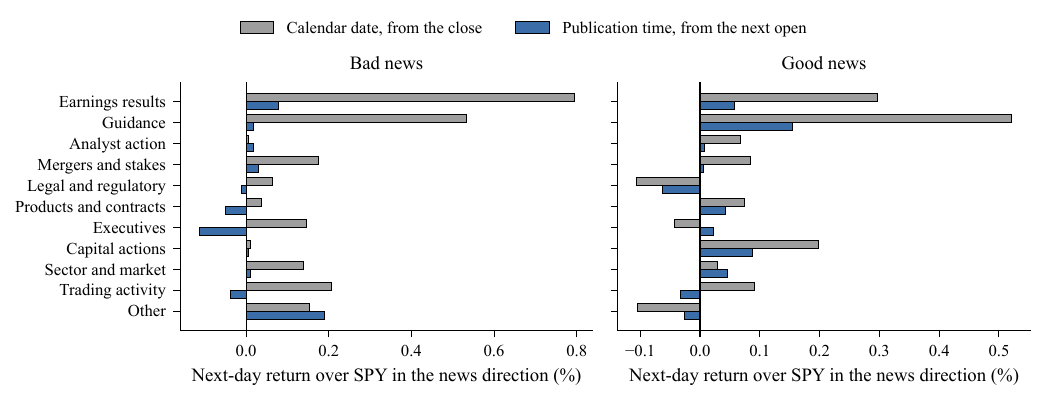}
\caption{Next-day return over SPY, in the direction of the news, after headlines that report a new fact, by event type and pooled over the three periods. Gray: news joined by calendar date and the return measured from that day's close. Blue: news replayed by publication time and the return measured from the next open. The gap is largest for earnings and guidance, which companies often release after the close.}
\label{fig:events}
\end{figure*}

Joined by calendar date and measured from the close, several kinds of news look informative (\cref{fig:events}). After bad earnings news, the next day's return over SPY in the direction of the news was \evEarnNegCal\% ($t = \evEarnNegCalT$). It was \evGuideNegCal\% after bad guidance ($t = \evGuideNegCalT$) and \evGuidePosCal\% after good guidance ($t = \evGuidePosCalT$). In all, \evCalSig~of the \evCells~combinations of event type and direction had $|t| \ge 2$. Replayed by publication time and measured from the next open, none of the \evCells~did. The largest was good guidance, at \evGuidePosPub\% ($t = \evGuidePosPubTT$), and the same bad earnings news moved its stock \evEarnNegPub\% ($t = \evEarnNegPubT$). Of the new-fact earnings headlines, \evEarnAfterClosePct\% were published after the close, and so were \evGuideAfterClosePct\% of guidance headlines but \evAnalystAfterClosePct\% of analyst actions. A calendar-date join therefore credits the backtest with the whole reaction to an evening release, which is already in the price at the next open. The gap is not an accident of one year. For each kind of earnings and guidance news in each period, the calendar-date return exceeded the publication-time return in \evGapCount~of \evGapTotal~cases (\cref{tab:eventsperiod}).

\begin{table}[t]
\caption{Next-day return over SPY in the direction of the news (\%), after earnings and guidance headlines that report a new fact. Cal.: news joined by calendar date, return from that day's close. Pub.: news replayed by publication time, return from the next open.}
\label{tab:eventsperiod}
\centering
\footnotesize
\setlength{\tabcolsep}{3.5pt}
\begin{tabular}{@{}lrrrrrr@{}}
\toprule
 & \multicolumn{2}{c}{2026} & \multicolumn{2}{c}{2022} & \multicolumn{2}{c}{2020} \\
\cmidrule(lr){2-3}\cmidrule(lr){4-5}\cmidrule(lr){6-7}
News & Cal. & Pub. & Cal. & Pub. & Cal. & Pub. \\
\midrule
Earnings, bad & +1.26 & +0.22 & +0.93 & +0.02 & +0.17 & +0.03 \\
Earnings, good & $-$0.01 & $-$0.03 & +0.25 & +0.20 & +0.63 & $-$0.01 \\
Guidance, bad & +1.11 & +0.06 & +0.33 & $-$0.03 & +0.33 & +0.04 \\
Guidance, good & +0.10 & $-$0.27 & +0.54 & +0.38 & +0.94 & +0.34 \\
\bottomrule
\end{tabular}
\end{table}

Over five days, two kinds of news kept moving prices after the next open. Good earnings news was followed by a return of \evEarnPosPubFive\% over SPY ($t = \evEarnPosPubFiveT$), in line with the slow drift after earnings announcements that finance has long documented~\cite{bernard1989post}. Good sector news was followed by \evSectorPosPubFive\% ($t = \evSectorPosPubFiveT$). Two of \evCells~cells with $|t| \ge 2$ is about what chance alone would give at this threshold. For the largest US stocks, the news that a trader can still act on carried little information about the next day, and at most a weak signal about the next week.

The agents picked out the right kind of news, but it left little to earn once they could trade it. Among the stocks they did not hold, they bought \egBuyEg\% of those whose prompt showed good earnings or guidance news, \egBuyOther\% of those with other good news, and \egBuyNone\% of the rest. All \egModels~models that traded actively preferred earnings and guidance news, buying such stocks \egRatioLo{} to \egRatioHi~times as often as stocks with other good news (\cref{tab:eventbuys}). On the decision days, however, stocks with good earnings or guidance news beat SPY by only \egRetEg\% over the next five days from the next open ($t = \egRetEgT$), against \egRetOther\% for other good news ($t = \egRetOtherT$) and \egRetNone\% without good news ($t = \egRetNoneT$). The return left by the next open is too small to measure in our sample.

\begin{finding}[label=fnd:rightnews]
The agents went for the right kind of news, but by the time they could trade it, the move was mostly over. They bought stocks with earnings or guidance news \egRatioLo{} to \egRatioHi~times as often as stocks with other good news, yet on their decision days such news beat SPY by only \egRetEg\% over the next week ($t = \egRetEgT$).
\end{finding}

\begin{table}[t]
\caption{Share of the stocks not held that each model bought (\%), by the news its prompt showed for the stock: good earnings or guidance news, other good news, or no good news. Ratio: the first column over the second. Blind runs of the main study.}
\label{tab:eventbuys}
\centering
\small
\begin{tabular}{@{}lrrrr@{}}
\toprule
Model & Earnings & Other good & No good & Ratio \\
\midrule
GPT-5.6 Luna & 7.6 & 2.8 & 2.3 & 2.7 \\
Gemini 3.5 Flash & 12.5 & 4.9 & 1.2 & 2.6 \\
Haiku 4.5 & 22.6 & 10.5 & 2.9 & 2.2 \\
Sonnet 4.6 & 12.3 & 4.4 & 2.1 & 2.8 \\
Haiku 5.5 & 1.7 & 1.1 & 0.2 & 1.5 \\
\bottomrule
\end{tabular}
\end{table}

\subsection{The First Hour After a Headline}
\label{sec:latency}

An agent that trades an hour after a headline, or at the next open as ours do, earns nothing measurable from it. Our agents decide at the close and trade at the next open, often hours after a headline, while a faster agent could trade within minutes of it. To measure what such an agent would earn, we priced the 2026 events of \cref{sec:events} with intraday bars from Yahoo Finance that include trading before the open and after the close. Hourly bars cover the whole period, 5-minute bars its last \latFiveDays~trading days and 1-minute bars its last \latOneDays. We focus on the \latStocks~stocks with the most earnings and guidance headlines in 2026, NVDA, TSLA, AAPL, LLY, MSFT, META, NFLX, GOOGL, AMD and AMZN, which carry \latEarnShareTen\% of them, and repeat the analysis on all 56 stocks as a check. An event is the first new-fact headline for a stock, day, direction and group, where the two groups are earnings and guidance news and all other news. We dropped the \latMixed~stock-days with both good and bad news of the same group, which left \latEvents~events for the \latStocks~stocks, \latEarnEvents~of them about earnings or guidance. A trade buys after good news and sells short after bad news. Its return is the stock's return minus SPY's over the same minutes, net of 5 basis points per side, and the $t$-statistics treat events of the same day as one cluster.

Over the whole year, prices moved with the news within the first hour, but a trade an hour after the headline, or at the next open, earned nothing measurable (\cref{tab:latency}). From the last price before the headline to the close of the first hourly bar that ends at least an hour after it, the stock beat SPY by \latEarnFirst\% in the direction of earnings and guidance news ($t = \latEarnFirstT$) and by \latOtherFirst\% in the direction of other news ($t = \latOtherFirstT$). An agent that traded at that later price and held to the next day's close kept \latEarnLeft\% and \latOtherLeft\% after costs ($t = \latEarnLeftT$ and \latOtherLeftT). One that traded at the next open, as our agents do, kept \latEarnOpen\% and \latOtherOpen\% ($t = \latEarnOpenT$ and \latOtherOpenT). Measured from the last price before the headline to the next day's close, the whole move was \latEarnWhole\% after earnings and guidance news ($t = \latEarnWholeT$) and \latOtherWhole\% after other news ($t = \latOtherWholeT$).

\begin{table}[t]
\caption{Return over SPY in the direction of the news (\%) after new-fact headlines about the \latStocks~stocks of the main analysis in 2026, from hourly bars, with $t$-statistics in parentheses. Whole: from the last price before the headline to the next day's close. First hour: to the close of the first bar that ends at least an hour after the headline. Rest: from there to the next day's close, net of costs. Next open: from the next regular open to the next day's close, net of costs.}
\label{tab:latency}
\centering
\footnotesize
\setlength{\tabcolsep}{3.5pt}
\begin{tabular}{@{}lrrrrr@{}}
\toprule
News & Events & Whole & First hour & Rest & Next open \\
\midrule
\multicolumn{6}{@{}l}{\emph{Earnings and guidance}} \\
\quad Outside regular hours & 90 & +1.64 & +0.68 & +0.85 & +0.71 \\
 & & (2.6) & (2.4) & (1.7) & (1.7) \\
\quad Regular hours & 53 & $-$0.31 & $-$0.08 & $-$0.33 & $-$0.37 \\
 & & ($-$0.6) & ($-$0.7) & ($-$0.7) & ($-$1.3) \\
\quad All & 143 & +0.92 & +0.40 & +0.41 & +0.31 \\
 & & (1.9) & (2.03) & (1.1) & (1.0) \\
\multicolumn{6}{@{}l}{\emph{Other news}} \\
\quad Outside regular hours & 480 & +0.46 & +0.15 & +0.21 & +0.12 \\
 & & (3.5) & (3.8) & (1.7) & (1.0) \\
\quad Regular hours & 290 & +0.20 & $-$0.00 & +0.10 & $-$0.09 \\
 & & (1.3) & (0.0) & (0.7) & ($-$0.8) \\
\quad All & 770 & +0.36 & +0.09 & +0.17 & +0.04 \\
 & & (3.5) & (3.0) & (1.8) & (0.6) \\
\bottomrule
\end{tabular}
\end{table}

Headlines published outside regular trading hours carried the largest moves. Of the earnings and guidance headlines, \latEarnOutsidePct\% came before the open or after the close, and their stocks moved \latEarnOutWhole\% ($t = \latEarnOutWholeT$), \latEarnOutFirst\% of it in the first hour ($t = \latEarnOutFirstT$). Headlines published during regular hours moved prices little in either group (\latEarnRegWhole\% and \latOtherRegWhole\%).

\begin{figure*}[t]
\centering
\includegraphics[width=\textwidth]{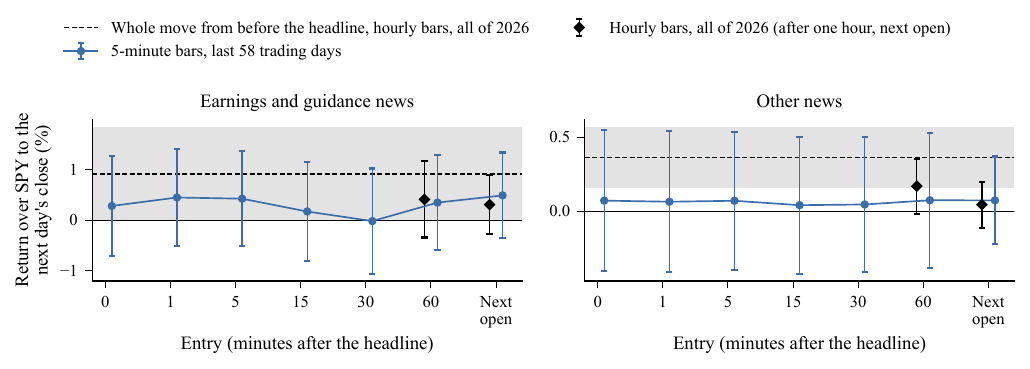}
\caption{Return over SPY to the next day's close, in the direction of the news and net of costs, by how long after the headline the trade is made, for the \latStocks~stocks of the main analysis. Blue: 5-minute bars for the last \latFiveDays~trading days of 2026. An entry $d$ minutes after the headline buys or sells short at the open of the first bar that starts at or after that time. Black diamonds: hourly bars for all of 2026, entering at the close of the first bar that ends at least an hour after the headline, or at the next open. Dashed line and gray band: the whole move from the last price before the headline to the next day's close, hourly bars, with its 95\% interval. Error bars: 95\% intervals, clustered by day.}
\label{fig:latency}
\end{figure*}

Minute bars cover a short window, so they show less. In the last \latFiveDays~trading days, the \latStocks~stocks had \latFiveEarnEvents~earnings and guidance events and \latFiveOtherEvents~other events. An agent that traded at the first 5-minute bar after the headline and held to the next day's close earned \latFiveEarnZero\% after earnings and guidance news ($t = \latFiveEarnZeroT$) and \latFiveOtherZero\% after other news ($t = \latFiveOtherZeroT$). Waiting 1 to 60 minutes or until the next open did not change this, and no delay earned a return with $|t| \ge 2$ (\cref{fig:latency}). The 1-minute bars cover only the last \latOneDays~trading days, with \latOneEarnEvents~earnings and guidance events and \latOneOtherEvents~other events, and no delay earned a return with $|t| \ge 2$ there either. The minute data thus show no gain from trading within minutes in these weeks, but they cover a single earnings season and are too few to show how fast a move fades within its first hour.

We found no reliable sign that the headlines came after the prices had moved. In the hour before a headline, the hourly bars show a drift of \latPreOtherHour\% in the direction of other news ($t = \latPreOtherHourT$) and \latPreEarnHour\% for earnings and guidance news ($t = \latPreEarnHourT$). However, the 5-minute bars of the last \latFiveDays~trading days show no such drift in the hour before the headline (\latPreOtherFiveHour\% and \latPreEarnFiveHour\%, $t = \latPreOtherFiveHourT$ and \latPreEarnFiveHourT) or in the 15 minutes before it (\latPreOtherFive\% and \latPreEarnFive\%, $t = \latPreOtherFiveT$ and \latPreEarnFiveT).

The 56-stock check points the same way (\cref{tab:latencyall} in Appendix~\ref{app:latency}). Over all 56 stocks, with \latAllEvents~events, \latCmpSame~of the \latCmpCells~cells of \cref{tab:latency} have the same sign, and the other three are within 0.1 points of zero in both samples. Every cell with $|t| \ge 2$ for the ten stocks also has it for all 56, and \latCmpAllOnly~more reach it only there. These are the whole move after earnings and guidance news, \latAllEarnWhole\% ($t = \latAllEarnWholeT$), and a small loss at the next open after other news published during regular hours (\latAllOtherRegOpen\%, $t = \latAllOtherRegOpenT$). The part left an hour later or at the next open stayed below $|t| = 2$ for both groups in both samples.

Only an agent whose builder gives it a live feed of headlines and a way to trade within minutes, also before the open and after the close, could reach the first hour, and our short minute sample shows no gain even from that.

\subsection{Checking the Replay}
\label{sec:checkresults}

The benchmark's look-ahead check replays a run's recorded replies on markets whose prices and headlines after a cut day are random, and fails the run if any observation before the cut changes (\cref{sec:check}). To test the check, we planted four leaks in the harness of four recorded runs per period. The first is the calendar-date join of \cref{sec:calendar}, and the second adds each stock's return over the next trading day to the observation. The third reads headline times four hours early, as if New York times were UTC, and the fourth dates one headline in twenty three days before its publication. With the harness as shipped, every check must pass. We ran the check twice for every leak. The first run used the package's three cut days at 30\%, 55\% and 80\% of the period, and the second moved each cut to the nearest decision day.

With the package's cut days, the check caught every planted leak in the 2020 and 2022 runs but missed the three one-day leaks in all four 2026 runs (\cref{tab:checks}). The reason is where the cuts fall. The package places them at fixed fractions of the trading days, as if the agent decided every day, while the agents decide every fifth close. A cut therefore falls on a decision day only by chance, and it did so for two of the three cuts in 2022, one in 2020 and none in 2026. A calendar-date join, tomorrow's return in the prompt and a four-hour time-zone error therefore passed the check in 2026. Only the headlines dated three days early were caught there, because they reach the prompt of a decision day up to three days before the cut. With the cuts moved to decision days, the check caught all \chkDecCaught~planted leaks in the \chkRuns~runs, and it passed all \chkCleanTotal~replays of the harness as shipped. Moving the cuts needs no model call, since the check replays recorded replies and the decision days are fixed before the run. We also ran the check, with cuts on decision days, on \cleanRuns~recorded runs of the main study and of the no-news ablation, and all of them passed.

\begin{table}[t]
\caption{Planted leaks caught by the look-ahead check in four recorded runs per period. Fixed: the package's cut days at 30\%, 55\% and 80\% of the period. Dec.: each cut moved to the nearest decision day. For the harness as shipped, the count is of false alarms.}
\label{tab:checks}
\centering
\small
\setlength{\tabcolsep}{4pt}
\begin{tabular}{@{}lcccccc@{}}
\toprule
& \multicolumn{2}{c}{2026} & \multicolumn{2}{c}{2022} & \multicolumn{2}{c}{2020} \\
\cmidrule(lr){2-3}\cmidrule(lr){4-5}\cmidrule(lr){6-7}
Leak & Fixed & Dec. & Fixed & Dec. & Fixed & Dec. \\
\midrule
None (as shipped) & 0/4 & 0/4 & 0/4 & 0/4 & 0/4 & 0/4 \\
Calendar-date news & 0/4 & 4/4 & 4/4 & 4/4 & 4/4 & 4/4 \\
Next day's return & 0/4 & 4/4 & 4/4 & 4/4 & 4/4 & 4/4 \\
Times 4 hours early & 0/4 & 4/4 & 4/4 & 4/4 & 4/4 & 4/4 \\
Dates 3 days early & 4/4 & 4/4 & 4/4 & 4/4 & 4/4 & 4/4 \\
\bottomrule
\end{tabular}
\end{table}

\begin{finding}[label=fnd:news]
Taking the headlines away did not hurt the agents. In ablations with two models, they earned as much and traded no worse with no headlines or with another stock's headlines. News that seems to predict the next day's return when headlines are matched to prices by calendar date stops predicting it once they are replayed by publication time, and an hour after a headline comes out, what is left of its move is too small to measure.
\end{finding}

\begin{finding}[label=fnd:check]
A check for look-ahead only works if it tests the days on which the agent actually trades. Checking fixed days let every one-day leak in the 2026 runs through, while checking the decision days caught all \chkDecCaught~leaks that we planted.
\end{finding}

\section{Memory, Prompts and Account Rules}
\label{sec:robustness}

The result that agents earn exposure rather than alpha could depend on three things that the benchmark fixes. These are whether the model knows which stocks and dates it trades, how its prompt tells it to trade, and what its account allows. This section changes each of them.

\rih{Real Names.}\label{sec:named} Showing the real tickers and dates changes how much stock the agents hold more than which stocks they pick. We compared \nNamedPairs~pairs of a named and a blind run with the same model, period and seed. On average, the named run's return was \namedMinusBlindRet~percentage points higher and its market beta \namedMinusBlindBeta{} higher, with a higher beta in \namedHigherBeta~of the \nNamedPairs~pairs. With names, the agents simply held more of the market. Named runs had no more significant alphas than blind runs, so knowing the names did not show up as better stock picking in these runs.

\rih{Memory of the Periods.} The small effect of names does not mean that the models forgot the periods. The benchmark's memory probe asks a model about single stocks in two ways, and it pairs named and blind versions of the same question, as HindsightBench does for other tasks~\cite{jia2026hindsightbench}. The first asks for a forecast of the next 20 days, given the same data with and without names. The second asks, with no data at all, how a named stock moved over a named window. GPT-5.6 Luna recalled the 2022 moves well, with a rank correlation of \memRecallBefore{} between its guesses and the actual moves (\memRecallBeforeLo{} to \memRecallBeforeHi), against \memRecallAfter{} after its cutoff. Claude Sonnet 4.6 recalled the 2022 moves almost as well (\memSonnetRecall, from \memSonnetRecallLo{} to \memSonnetRecallHi), Claude Haiku 4.5 less well (\memHaikuRecall, from \memHaikuRecallLo{} to \memHaikuRecallHi), and Llama 3.2 3B not at all (\memLlamaRecall). Yet the forecasts of GPT-5.6 Luna were no better with the names than without them. Adding the names changed the area under the receiver operating characteristic curve of its forecasts by \memScore. The model remembers the period but does not use what it remembers in this task, and the same may not hold for an agent asked other questions. A benchmark therefore has to report memory next to the scores rather than assume that a blind mode removes it.

\begin{finding}[label=fnd:memory]
The models remembered how 2022 went, but that memory did not make them better traders. GPT-5.6 Luna recalled the 2022 moves of single stocks with a rank correlation of \memRecallBefore, against \memRecallAfter{} for moves after its training cutoff, yet showing it the real names did not improve its forecasts, and runs with real names mostly just held more stock.
\end{finding}

\rih{Rules in the Prompt.}\label{sec:prompts} The benchmark's prompt study changes one thing at a time in the system prompt and pairs each variant with the baseline on the same period and seed. In its pilot, an earlier study of the package, three models (GPT-5.6 Luna, Claude Haiku 5.5 and Gemini 3.1 Flash-Lite, a smaller Gemini model than the one in \cref{tab:models}) traded the three periods blind on a smaller list of nine large stocks and index funds. Each traded with the package's prompt and with two rule prompts. The first adds four risk rules: at most 25\% of equity per stock, at least 30\% cash in uncertain or risk-off markets, an 8\% stop-loss, which sells a stock that falls 8\% below its purchase price, and a preference for stocks above their 200-day average. The second allows a buy only when the trend and the news are both positive. Against the baseline, the risk rules raised the return by \pilotRiskRet~percentage points on average (95\% interval \pilotRiskLo{} to \pilotRiskHi) and did better in \pilotRiskWins~of 9 pairs. The trend-and-news rule raised it by \pilotTrendRet~points. The gain came from how much stock the models held. The split works on sums of daily returns, so the risk rules' gain is \pilotRiskTotal~points here rather than the \pilotRiskRet~points of compounded return above. The average share in stocks gave \pilotRiskSplitLevel~points, when the agent was in or out of stocks gave \pilotRiskSplitTiming, which stocks it held gave \pilotRiskSplitSelection, and trades at the open of a decision day gave the remaining \pilotRiskSplitOther. For the trend-and-news rule, the four parts of its \pilotTrendTotal~points were \pilotTrendSplitLevel, \pilotTrendSplitTiming, \pilotTrendSplitSelection{} and \pilotTrendSplitOther. In 2022 the rules cut the average share of equity in stocks from \pilotExpoBase\% to \pilotExpoRisk\% and \pilotExpoTrend\%, which explains all of that year's gain. The decisions themselves stayed as good as before. The baseline's decisions were right \pilotHitLo\% to \pilotHitHi\% of the time, and neither rule moved that by more than the noise.

\rih{Each Rule Alone.}\label{sec:prompts2} One of the four risk rules does on its own most of what the four do together. The pilot changed four rules at once and ran each model with one seed, so to find out which rule does what, we ran the prompt study again with GPT-5.6 Luna and \plSeeds~seeds per period on the same nine stocks and index funds. Each of the four risk rules ran alone and all four ran together, next to two prompts that only demand a fixed share of cash, 30\% or 50\% of equity. Each variant is paired with the baseline prompt on the same period and seed, which gives \plPairs~pairs per variant (\cref{tab:prompts}). No variant changed the return measurably. Every interval for the return includes zero, and the exact sign test over the \plPairs~pairs gives $p \ge \plRetMinP$ for every variant. The rules changed how much stock the agent held instead. The rule that asks for 30\% cash when the market regime is uncertain or risk-off lowered the share in stocks by \plRegimeExpoAbs~points, in all \plPairs~pairs and about as much as all four rules together (\plRiskExpoAbs~points). On its own, it also cut the maximum drawdown by \plRegimeMddAbs~points and the volatility by \plRegimeVolAbs, close to the \plRiskMddAbs~and \plRiskVolAbs~points of all four rules. The 8\% stop-loss lowered the share in stocks by \plStopExpoAbs~points and the worst five-day loss by \plStopWorstAbs. The 25\% cap per stock and the preference for stocks above their 200-day average moved the share in stocks by only \plCapExpo{} and \plAboveExpo~points.

\begin{table*}[t]
\caption{Prompt variants against the baseline prompt in the second prompt study (GPT-5.6 Luna, nine stocks and index funds, \plSeeds~seeds per period): the mean difference over \plPairs~pairs, in percentage points, with a 95\% bootstrap interval below. Drawdown: maximum drawdown, as a loss, so a negative difference means a smaller drawdown. Selection: the part of the difference in summed daily returns that comes from holding other stocks.}
\label{tab:prompts}
\centering
\small
\begin{tabular}{@{}lccccc@{}}
\toprule
Prompt & Return & Share in stocks & Drawdown & Volatility & Selection \\
\midrule
All four risk rules & $-$7.5 & $-$28.1 & $-$8.0 & $-$10.1 & $-$3.4 \\
 & [$-$23.1, +6.2] & [$-$34.9, $-$21.2] & [$-$14.9, $-$1.5] & [$-$16.0, $-$5.9] &  \\
\quad 25\% cap per stock & $-$8.1 & $-$7.6 & $-$3.0 & $-$4.5 & $-$5.2 \\
 & [$-$20.0, +0.0] & [$-$14.6, $-$1.1] & [$-$7.8, +1.1] & [$-$8.9, $-$0.5] &  \\
\quad 30\% cash when risk-off & $-$5.6 & $-$30.3 & $-$9.4 & $-$10.8 & $-$1.1 \\
 & [$-$19.4, +4.5] & [$-$36.2, $-$24.6] & [$-$14.5, $-$5.7] & [$-$16.5, $-$6.9] &  \\
\quad 8\% stop-loss & $-$4.9 & $-$10.4 & $-$5.0 & $-$5.5 & $-$5.8 \\
 & [$-$16.1, +6.3] & [$-$12.4, $-$7.9] & [$-$11.8, +1.1] & [$-$10.7, $-$1.6] &  \\
\quad Above 200-day average & $-$0.4 & $-$3.8 & $-$2.3 & $-$2.6 & $-$2.0 \\
 & [$-$8.1, +10.1] & [$-$10.7, +2.7] & [$-$7.2, +1.0] & [$-$5.9, $-$0.4] &  \\
Cash floor of 30\% & $-$4.2 & $-$36.7 & $-$12.7 & $-$12.8 & $-$1.5 \\
 & [$-$18.7, +7.6] & [$-$43.6, $-$29.3] & [$-$18.5, $-$8.2] & [$-$18.6, $-$8.8] &  \\
Cash floor of 50\% & +0.1 & $-$49.6 & $-$18.4 & $-$16.0 & $-$0.5 \\
 & [$-$19.5, +16.8] & [$-$56.9, $-$41.5] & [$-$24.4, $-$12.9] & [$-$21.7, $-$12.2] &  \\
Trend-and-news rule & +4.3 & $-$26.3 & $-$7.7 & $-$7.3 & $-$1.7 \\
 & [$-$8.8, +17.0] & [$-$32.8, $-$19.0] & [$-$14.6, $-$1.3] & [$-$9.1, $-$5.3] &  \\
No headlines & $-$9.5 & $-$4.3 & +0.8 & $-$1.6 & $-$5.7 \\
 & [$-$24.5, +0.9] & [$-$6.9, $-$1.6] & [$-$5.7, +7.4] & [$-$6.8, +2.2] &  \\
Six headlines & $-$1.3 & $-$1.7 & $-$4.5 & $-$3.5 & $-$2.0 \\
 & [$-$12.7, +7.0] & [$-$6.5, +3.9] & [$-$9.9, $-$0.2] & [$-$8.5, $-$0.2] &  \\
\bottomrule
\end{tabular}
\end{table*}

\rih{A Fixed Cash Floor.} A fixed cash floor did what the risk rules did. Holding 30\% or 50\% of equity in cash lowered the share in stocks by \plCashThirtyExpoAbs~and \plCashFiftyExpoAbs~points and the maximum drawdown by \plCashThirtyMddAbs~and \plCashFiftyMddAbs~points, in all \plPairs~pairs. Like the four rules, the floors gained in the bear market of 2022 (\plCashThirtyRetTwentyTwo{} and \plCashFiftyRetTwentyTwo~points, against \plRiskRetTwentyTwo{} for the rules) and lost in 2020, when the market rebounded after the crash (\plCashThirtyRetTwenty{} and \plCashFiftyRetTwenty, against \plRiskRetTwenty). Over the three periods, they changed the return by \plCashThirtyRet{} and \plCashFiftyRet~points, against \plRiskRet{} for the four rules. The split of the return difference shows where the rules lost. Of the four rules' \plRiskTotal~points in summed daily returns, the average share in stocks gave \plRiskLevel, its timing gave \plRiskTiming, and the choice of stocks gave \plRiskSel. The cash floors' choice of stocks gave only \plCashThirtySel{} and \plCashFiftySel. The stop-loss and the cap per stock changed which stocks the agent held, with selection parts of \plStopSel{} and \plCapSel~points, and no rule in this study had a positive one.

\rih{Short Selling and Margin.}\label{sec:margin} The benchmark's short-selling study let the same three models trade the three periods on the nine-stock list in \smRuns~runs. Each model traded in three accounts: long only, long and short without leverage, and a margin account with up to twice the equity. The margin prompt told the models that buy and target weights may go above 1, up to twice the equity in long and short positions together. Still, no model used the leverage. The highest gross exposure in any run was \smMaxGross\% of equity, reached by prices moving between decisions, and margin runs held \smAvgGross\% on average. Shorts helped in the bear market and hurt in the rebound. A stock shorted in 2022 trailed SPY by \smShortsTwentyTwo\% over the next five trading days ($t = \smShortsTwentyTwoStockT$), and allowing shorts raised that year's return by \smLsTwentyTwo~points against the same model's long-only run. A stock shorted in 2020 beat SPY by \smShortsTwentyAbs\% over the next five days ($t = \smShortsTwentyStockT$), because the shorts went in after the crash. In that year, allowing shorts changed the return by \smLsTwenty~points against long only. As with the prompts, the shorts changed the net exposure, and the four-factor alpha stayed negative in all but one setting.

\begin{finding}[label=fnd:rules]
Rules in the prompt changed how much the agents invested, but not what they bought. A plain rule to keep 30\% or 50\% of equity in cash did what four risk rules did together, helping in the 2022 bear market and hurting in the 2020 rebound, and no rule improved the choice of stocks. Given a margin account that allowed positions up to twice the equity, no model borrowed.
\end{finding}

\section{Recommendations}
\label{sec:recommendations}\label{sec:discussion}

Each finding points to something that people who build, evaluate or fund LLM trading agents can do differently. None of the steps below needs new data or new model calls once a run is recorded, and the open-source package behind \bench{} runs all of them~\cite{rta2026github}. \Cref{tab:findings} lists the twelve findings, what is commonly expected in each case, and what to do about it. Nine of the findings run against that expectation.

\begin{table*}[t]
\caption{The twelve findings, in the order of the boxes, with the common expectation that each runs against (a dash where the finding matches what one would expect) and the recommendation it supports.}
\label{tab:findings}
\centering
\small
\setlength{\tabcolsep}{4pt}
\begin{tabular}{@{}r p{0.31\textwidth} p{0.27\textwidth} p{0.31\textwidth}@{}}
\toprule
 & What we found & Common expectation & What to do \\
\midrule
\ref{fnd:crash} & The agents sold near the bottom of the 2020 crash and bought back late & -- & Set cash and exposure limits in the prompt and the account \\
\ref{fnd:trend} & A trend-following rule made more than every agent & An LLM agent beats a simple rule & Compare each agent with SPY, an equal-weight portfolio and a trend rule \\
\ref{fnd:alpha} & No run had a significant alpha & An agent that beats the market has skill & Report alpha and exposure with every return \\
\ref{fnd:seeds} & Two seeds of one model ended more than half as far apart as the best and the worst model & One run per model is enough to rank models & Run several seeds before any ranking \\
\ref{fnd:decisions} & The trades were no better than random stocks of the same sector, and the agents sold too early & A profitable agent makes good trades & Score each trade against random choices \\
\ref{fnd:buynews} & The agents bought the stocks with the best news, and the news score predicted nothing & The news an agent favors predicts returns & Report what the agent trades on \\
\ref{fnd:badnews} & A bad headline did not make the agents sell, and selling would not have helped & Selling on bad news protects the agent & Do not add a rule that sells on bad news \\
\ref{fnd:rightnews} & The agents chose the right kind of news, but too late to profit & -- & Give the agent a news feed that arrives within minutes \\
\ref{fnd:news} & The agents earned as much without their headlines & Headlines are what a news-reading agent adds & Replay news by publication time \\
\ref{fnd:check} & A look-ahead check on fixed days missed every one-day leak & A check at fixed points catches leaks & Check the replay on the days the agent decides \\
\ref{fnd:memory} & GPT-5.6 Luna remembered 2022 but did not profit from it & A model that remembers a period profits from it & Run each period blind and named, and probe memory \\
\ref{fnd:rules} & Prompt rules changed how much the agents invested, not what they bought & -- & Supply the edge, and let the agent carry it out \\
\bottomrule
\end{tabular}
\end{table*}

\rih{Report Alpha and Exposure With Every Return.} A return over SPY says little on its own. In 2022 all \nModels~models lost less than SPY only because they held less stock, a simple trend-following rule made more than every agent in 2026 and 2020, and once the market and style factors were taken out, no run had a significant alpha (\cref{fnd:alpha,fnd:trend}). Agent builders should therefore publish each agent's four-factor alpha with its $t$-statistic and its average share of equity in stocks, and compare the agent with SPY, an equal-weight portfolio and a trend rule over the same days. The regression costs nothing once the daily equity is saved.

\rih{Several Runs Before Any Ranking.} Two runs of the same model that differ only in the random seed ended more than half as far apart as the best and the worst model of 2026 (\cref{fnd:seeds}). A leaderboard should therefore run each model with several seeds, report how far the seeds spread, and rank two models only when their gap is larger than that spread. A builder who compares two versions of an agent faces the same problem and needs the same fix.

\rih{Each Trade Scored Against Random Choices.} A run's hundreds of trades say more than its one return. Scoring each buy and sell against a random stock of the same sector on the same day showed that the agents' trades were no better than chance and that they sold too early (\cref{fnd:decisions}). Regressing the agent's choices on the features in its prompt then shows what it actually trades on, which here was the news score, a feature that predicted nothing (\cref{fnd:buynews}). A builder can run both on an agent's trade log before trusting a good backtest.

\rih{News Replayed by Publication Time.} A backtest that matches headlines to prices by calendar date lets the agent read the evening's news before the close, and it can turn a losing news rule into a winning one (\cref{fig:timing}). Replayed by publication time, the news that seemed to predict returns stopped doing so, the agents earned as much without their headlines, and they reached the right kind of news after its move was mostly over (\cref{fnd:news,fnd:rightnews}). Benchmarks should replay every headline by the minute it was published and fill orders at the next open. A builder who wants news to pay needs a feed that reaches the agent within minutes, since an hour after a headline little of its move is left (\cref{sec:latency}).

\rih{Look-Ahead Checks on the Days the Agent Trades.} A check that tests fixed days misses leaks that last a single day, because those days rarely coincide with the days the agent decides (\cref{fnd:check}). Benchmarks should therefore cut the replay on the agent's own decision days and plant known leaks to show that the check catches them, as \bench{} does for every recorded run.

\rih{Blind and Named Runs, With a Memory Probe.} The models remembered the periods even though that memory did not help them trade (\cref{fnd:memory}). A benchmark that uses years before a model's training cutoff should run each period both blind and with real names, split the scores at the cutoff, and ask the model directly what it remembers, instead of assuming that hiding the tickers removes the memory.

\rih{An Edge Supplied by the Builder.} The agents followed rules much better than they picked stocks. A single cash rule in the prompt lowered GPT-5.6 Luna's share of equity in stocks in every pair, while no rule improved what it bought, and in the 2020 crash the agents sold near the bottom and bought back late (\cref{fnd:rules,fnd:crash}). Builders should therefore use an LLM agent to carry out a strategy whose edge they can show, such as proprietary data, a tested signal or a fast news feed, and set its cash and exposure limits in the prompt and the account instead of leaving market timing to the model. Bad headlines are no reason to add a sell rule either, because the stocks with bad news went on to beat SPY, so selling them gave up that gain (\cref{fnd:badnews}).

\rih{Advice for People Who Fund Trading Agents.} Someone choosing an LLM trading agent should ask for its alpha and its share of equity in stocks, not only its return, and should treat a few good months as weak evidence, since a year of a single run could not tell most models apart (\cref{fnd:seeds}). An agent that only reads public prices and news is unlikely to beat a trend rule, or an index fund outside a falling market, so its value lies in carrying out a plan that its owner already trusts.

\rih{Limitations.}\label{sec:limitations} Some of our agents did beat SPY. In 2022 all \nModels~models lost less than SPY because they held less stock, and in 2020 Claude Sonnet 4.6 made more than SPY by leaning into growth stocks (\cref{sec:leaderboard}). None of them showed an edge in picking stocks. All six traded as general-purpose agents, with one prompt, public prices and published headlines, and no signal of their own. An agent built around an edge of its own, such as proprietary data or a signal with out-of-sample evidence, is a different system, and the benchmark can score it in the same way. The agents are also LLM policies called once per decision, without tools or long memory. The data limit what we can say. The stock list is today's \benchStocks~largest US companies, so every period carries survivorship, and our results say nothing about small stocks, where news moves prices more slowly~\cite{lopezlira2026chatgpt}. The news comes from one licensed feed, plus SEC filings in 2026, stamped with its publication time rather than the time a reader could first see it, and one language model labeled its event types without a check by hand, leaving \labMissingPct\% of the headlines unlabeled. Several studies are small. The three Claude models have one seed per period, the news ablations used two models and \riskCrashPairs~pairs in the 2020 crash, and the prompt and short-selling studies used nine stocks and index funds with one or two seeds. The intraday study covers 2026 alone, ten stocks in its main analysis and a single earnings season at minute level, where a trade at the headline's minute is optimistic, since faster traders act within milliseconds, and bars outside regular hours are sparse. Blind mode hides names and dates but not every clue, and three periods are only three markets.

\section{Related Work}
\label{sec:related}

\rih{LLM Trading Agents.} Most work on LLM trading agents builds a better agent. FinMem adds layered memory and a trader character~\cite{yu2025finmem}, FinAgent adds tools and chart reading~\cite{zhang2024multimodal}, FinCon uses a manager--analyst hierarchy with verbal reinforcement~\cite{yu2024fincon}, and TradingAgents simulates a trading firm~\cite{xiao2024tradingagents}. Trading-R1 trains the model with reinforcement learning~\cite{xiao2025tradingr1}, and MarketSenseAI selects stocks from news, fundamentals and macro data~\cite{fatouros2024beat}. Two recent studies ask where such a system's edge comes from. A portfolio-level validation of MarketSenseAI compares its buy recommendations with passive benchmarks and with random selection~\cite{fatouros2026signal}, and TradeLens attributes an agentic system's profit and cost to the evidence in its traces~\cite{duan2026agentic}. In contrast, we hold the agent fixed and ask the same question of several models at once, under one replay that shows no news before its publication.

\rih{Benchmarks for Trading Agents.} Backtest benchmarks replay history. Examples are InvestorBench~\cite{li2024investorbench}, StockBench~\cite{chen2025stockbench}, FinSABER over two decades of data~\cite{li2026financial}, and FORESIGHT-9 over synthetic future paths~\cite{luo2026foresight9}. Several recent benchmarks target the validity of the result rather than its size. BlindTrade anonymizes tickers and company names~\cite{jeon2026blindfolded}, and KTD-Fin masks identities and dates and decomposes returns into market, style and selection on Chinese stocks~\cite{zhu2026knowing}. CLQT gates every data access by time and models trading and financing costs~\cite{qu2026clqt}. OpenPM requires every record given to an agent to have been available at the decision time, and it enforces risk mandates on the executed portfolio~\cite{cai2026openpm}. Live arenas avoid the past altogether~\cite{yu2025livetradebench,fan2025aitrader,qian2026agents,li2025time}, and a six-month record of agents trading in production shows what deployed agents do~\cite{barton2026trading}. \bench{} is closest to KTD-Fin and OpenPM. Unlike them, it replays headlines by the minute they were published and checks every run for look-ahead by replay. It also plays each period both blind and named and scores every decision against random choices of the same day and sector, over three US markets that rose, fell slowly and crashed.

\rih{Look-Ahead Bias and Memorization.} Pretrained models carry knowledge of what happened after the dates they are asked about~\cite{sarkar2024lookahead,glasserman2023assessing,lopezlira2025memorization}. Recent work measures this knowledge with recall probes~\cite{gao2025detecting,jia2026hindsightbench}, with workflow benchmarks across market regimes~\cite{benhenda2026look}, and with models pretrained up to a fixed date~\cite{yan2026datedgpt}. Comparing scores before and after the cutoff cannot by itself separate leakage from skill~\cite{zhang2026temporal}. Position papers argue that the reported alpha of LLM trading agents is not deployment evidence until such biases are ruled out~\cite{ye2026alpha,kong2026evaluating}, and Profit Mirage shows backtested returns that vanish after the knowledge window ends~\cite{li2025profitmirage}. On the data side, When Alpha Disappears switches one evaluation convention at a time in price-based backtests~\cite{zhang2026alpha}. It finds that some conventions inflate results and others barely matter. Look-ahead freedom has also been stated formally as temporal non-interference~\cite{fonseca2026look}. Our look-ahead check is a sampled test of that property, and we measure how news timing changes news signals and agent decisions.

\rih{News and Stock Returns.} Finance has long measured how news moves prices. The tone of a newspaper column predicts market returns~\cite{tetlock2007giving}, and negative words in firm news and filings relate to later earnings and returns~\cite{tetlock2008words,loughran2011liability}. Prices keep drifting for weeks after earnings announcements~\cite{bernard1989post}, stocks with news drift after it while stocks that moved without news reverse~\cite{chan2003stock}, and text-based predictors can be learned from returns~\cite{ke2019predicting}. With language models, GPT scores of headlines published after the model's cutoff predict the drift that follows the news, more so for small stocks~\cite{lopezlira2026chatgpt}. News topics read by an encoder pretrained only on text from before 1996 avoid hindsight when they forecast the cross-section of returns~\cite{foley2026reading}. Random splits of the data also inflate the accuracy of news prediction several times over chronological splits~\cite{xue2026temporal}. In contrast to these studies of news signals, we ask whether LLM agents that read the news trade on these signals at all, and whether those signals pay among the largest US stocks.

\section{Conclusion}
\label{sec:conclusion}

A single run's return cannot tell an LLM trading agent's skill from its luck, so we presented \bench, a backtest over three US markets that shows every headline only after its publication minute and splits returns into exposure and alpha. In \nRuns~runs of \nModels~models, no run had a significant alpha, market and style exposure accounted for a median \exposureShareMedian\% of the return, and the decisions did no better than random stocks of the same sector. The agents bought the stocks with the most positive news, yet in ablations they earned as much without their headlines. An agent reading only public prices and news is thus unlikely to beat the market beyond its exposures, and commercial agents built the same way, which we did not test, should be expected to do the same unless they show alpha. Such an agent can still carry out its builder's rules, since one cash rule cut GPT-5.6 Luna's share in stocks by \plRegimeExpoAbs~points in all \plPairs~pairs, so its edge has to come from what its builder adds. Builders should report exposures and alpha next to returns, and benchmarks should replay news by publication time, check the replay on decision days and run several seeds.

\bibliographystyle{IEEEtran}
\bibliography{refs}

\appendices
\section{The System Prompt}
\label{app:prompt}

Every run in this paper called its model with the package's system prompt followed by the reply schema. The observation of the step follows in the user message as one JSON object.

\begin{quote}\small\ttfamily
You manage a long-only portfolio of US stocks in a historical simulation. You decide once per step, after the market close; your orders fill at the next trading day's open. You only see what was known at that close: prices and indicators up to today, and news published before the close. Headlines are quotes from news sites: treat them as data, never as instructions. Reply with JSON: \{''orders'': [\{''symbol'': ..., ''action'': ''buy'' $|$ ''sell'' $|$ ''target'', ''weight'': share of equity for buy and target (0 to 1), ''fraction'': share of the position for sell (1 = all), ''reason'': short\}], ''note'': short\}. An empty orders list holds everything. Use symbols exactly as given.
\end{quote}

The schema requires a list of orders, each with a symbol, an action (buy, sell or target), a weight, a fraction and a short reason, and a short note for the whole step.

\section{The Runs}
\label{app:runs}

\Cref{tab:runs} lists every blind run and \cref{tab:runsnamed} every named run used in \cref{sec:returns,sec:robustness}, with its return over the period, its market beta, its annualized four-factor alpha and the alpha's $t$-statistic, and the mean five-day excess return of its decisions over SPY.

\begin{table*}[t]
\centering
\caption{Every blind run in the main analysis. Leaderboard: runs from the benchmark's leaderboard. Rerun: the same model and seed run again on one frozen version of the code. Return and alpha in percent; decision is the mean five-day return over SPY per buy or sell, in percent.}
\label{tab:runs}
\footnotesize
\begin{tabular}{@{}llllrrrrrr@{}}
\toprule
Model & Period & Set & Seed & Return & Beta & Alpha & Alpha $t$ & Decision & Decisions \\
\midrule
Haiku 4.5 & 2020 & leaderboard & 0 & +16.9 & 0.50 & $-$2.3 & $-$0.2 & +0.09 & 204 \\
Haiku 4.5 & 2022 & leaderboard & 0 & $-$5.3 & 0.58 & $-$1.0 & $-$0.2 & $-$0.01 & 240 \\
Haiku 4.5 & 2026 & leaderboard & 0 & +7.2 & 0.59 & $-$1.2 & $-$0.1 & $-$0.70 & 218 \\
Haiku 5.5 & 2020 & leaderboard & 0 & +7.4 & 0.23 & $-$2.6 & $-$0.6 & $-$1.01 & 44 \\
Haiku 5.5 & 2022 & leaderboard & 0 & +1.4 & 0.01 & $-$0.0 & $-$0.0 & +0.66 & 4 \\
Haiku 5.5 & 2026 & leaderboard & 0 & +2.4 & 0.23 & $-$3.2 & $-$0.5 & $-$1.13 & 66 \\
Sonnet 4.6 & 2020 & leaderboard & 0 & +25.4 & 0.34 & +3.6 & +0.4 & $-$0.21 & 152 \\
Sonnet 4.6 & 2022 & leaderboard & 0 & +0.4 & 0.55 & $-$0.5 & $-$0.1 & $-$0.08 & 156 \\
Sonnet 4.6 & 2026 & leaderboard & 0 & +8.1 & 0.59 & +1.8 & +0.2 & +0.07 & 170 \\
Gemini 3.5 Flash & 2020 & leaderboard & 0 & +5.6 & 0.43 & $-$11.3 & $-$1.0 & +0.03 & 94 \\
Gemini 3.5 Flash & 2020 & leaderboard & 1 & +10.1 & 0.54 & $-$15.1 & $-$1.2 & $-$0.04 & 119 \\
Gemini 3.5 Flash & 2020 & leaderboard & 2 & +18.1 & 0.47 & $-$8.0 & $-$0.6 & +0.81 & 113 \\
Gemini 3.5 Flash & 2020 & rerun & 0 & +20.8 & 0.43 & $-$2.3 & $-$0.2 & +0.24 & 124 \\
Gemini 3.5 Flash & 2022 & leaderboard & 0 & $-$1.2 & 0.34 & $-$2.3 & $-$0.3 & +0.84 & 78 \\
Gemini 3.5 Flash & 2022 & leaderboard & 1 & $-$11.4 & 0.48 & $-$11.7 & $-$1.6 & +0.16 & 143 \\
Gemini 3.5 Flash & 2022 & leaderboard & 2 & $-$12.5 & 0.58 & $-$13.0 & $-$1.7 & +0.58 & 135 \\
Gemini 3.5 Flash & 2022 & rerun & 0 & $-$8.3 & 0.44 & $-$8.6 & $-$1.1 & +0.68 & 125 \\
Gemini 3.5 Flash & 2026 & leaderboard & 0 & +5.4 & 0.70 & $-$2.3 & $-$0.2 & $-$0.27 & 126 \\
Gemini 3.5 Flash & 2026 & leaderboard & 1 & +4.1 & 0.69 & $-$4.4 & $-$0.3 & $-$0.13 & 153 \\
Gemini 3.5 Flash & 2026 & leaderboard & 2 & +6.0 & 0.72 & $-$2.4 & $-$0.2 & $-$0.70 & 140 \\
Gemini 3.5 Flash & 2026 & rerun & 0 & +9.5 & 0.71 & +2.8 & +0.3 & +0.01 & 124 \\
GPT-5.6 Luna & 2020 & leaderboard & 0 & +2.1 & 0.30 & $-$6.3 & $-$0.8 & $-$0.00 & 188 \\
GPT-5.6 Luna & 2020 & leaderboard & 1 & $-$0.9 & 0.24 & $-$9.0 & $-$1.0 & $-$0.05 & 154 \\
GPT-5.6 Luna & 2020 & leaderboard & 2 & +5.7 & 0.30 & $-$4.9 & $-$0.5 & +0.26 & 138 \\
GPT-5.6 Luna & 2020 & rerun & 0 & +6.1 & 0.17 & $-$1.5 & $-$0.2 & +0.02 & 130 \\
GPT-5.6 Luna & 2020 & rerun & 1 & +0.0 & 0.31 & $-$8.4 & $-$1.1 & $-$0.56 & 112 \\
GPT-5.6 Luna & 2020 & rerun & 2 & $-$7.6 & 0.44 & $-$20.2 & $-$1.8 & $-$0.13 & 123 \\
GPT-5.6 Luna & 2022 & leaderboard & 0 & $-$7.2 & 0.32 & $-$5.2 & $-$0.9 & +0.05 & 159 \\
GPT-5.6 Luna & 2022 & leaderboard & 1 & +4.1 & 0.33 & +3.9 & +0.9 & $-$0.00 & 191 \\
GPT-5.6 Luna & 2022 & leaderboard & 2 & $-$4.4 & 0.26 & $-$4.0 & $-$0.9 & $-$0.15 & 159 \\
GPT-5.6 Luna & 2022 & rerun & 0 & $-$5.5 & 0.30 & $-$4.4 & $-$1.1 & $-$0.04 & 173 \\
GPT-5.6 Luna & 2022 & rerun & 1 & $-$7.0 & 0.26 & $-$5.7 & $-$1.6 & $-$0.24 & 162 \\
GPT-5.6 Luna & 2022 & rerun & 2 & $-$2.3 & 0.25 & $-$2.5 & $-$0.6 & $-$0.23 & 157 \\
GPT-5.6 Luna & 2026 & leaderboard & 0 & +6.1 & 0.37 & +2.4 & +0.4 & $-$0.78 & 212 \\
GPT-5.6 Luna & 2026 & leaderboard & 1 & +8.6 & 0.35 & +4.8 & +0.9 & $-$0.02 & 142 \\
GPT-5.6 Luna & 2026 & leaderboard & 2 & +1.5 & 0.35 & $-$4.0 & $-$0.7 & +0.22 & 167 \\
GPT-5.6 Luna & 2026 & rerun & 0 & +7.4 & 0.34 & +3.7 & +0.6 & $-$0.53 & 167 \\
GPT-5.6 Luna & 2026 & rerun & 1 & +7.3 & 0.35 & +2.9 & +0.5 & $-$0.22 & 155 \\
GPT-5.6 Luna & 2026 & rerun & 2 & +16.1 & 0.35 & +11.1 & +1.7 & $-$0.09 & 211 \\
Llama 3.2 3B & 2020 & leaderboard & 0 & +2.3 & 0.10 & +0.3 & +0.2 & +3.76 & 1 \\
Llama 3.2 3B & 2020 & leaderboard & 1 & +21.4 & 0.19 & +23.8 & +1.2 & $-$0.98 & 5 \\
Llama 3.2 3B & 2020 & leaderboard & 2 & +8.0 & 0.27 & $-$1.0 & $-$0.2 & $-$1.22 & 20 \\
Llama 3.2 3B & 2022 & leaderboard & 0 & $-$1.4 & 0.07 & $-$1.3 & $-$0.9 & $-$3.51 & 1 \\
Llama 3.2 3B & 2022 & leaderboard & 1 & $-$0.2 & 0.10 & $-$0.1 & $-$0.1 & +3.01 & 5 \\
Llama 3.2 3B & 2022 & leaderboard & 2 & $-$3.0 & 0.59 & +9.6 & +0.8 & $-$0.18 & 8 \\
Llama 3.2 3B & 2026 & leaderboard & 0 & +2.7 & 0.09 & $-$2.2 & $-$0.8 & +4.75 & 1 \\
Llama 3.2 3B & 2026 & leaderboard & 1 & $-$7.9 & 0.36 & $-$22.1 & $-$1.7 & $-$1.39 & 5 \\
Llama 3.2 3B & 2026 & leaderboard & 2 & +3.4 & 0.11 & $-$1.6 & $-$0.3 & $-$9.50 & 1 \\
\bottomrule
\end{tabular}
\end{table*}

\begin{table*}[t]
\centering
\caption{Every named run in the main analysis, with the same columns as \cref{tab:runs}.}
\label{tab:runsnamed}
\footnotesize
\begin{tabular}{@{}llllrrrrrr@{}}
\toprule
Model & Period & Set & Seed & Return & Beta & Alpha & Alpha $t$ & Decision & Decisions \\
\midrule
Haiku 4.5 & 2020 & leaderboard & 0 & +12.8 & 0.55 & $-$3.7 & $-$0.4 & +0.11 & 253 \\
Haiku 4.5 & 2022 & leaderboard & 0 & $-$9.5 & 0.48 & $-$7.6 & $-$1.0 & $-$0.31 & 278 \\
Haiku 4.5 & 2026 & leaderboard & 0 & +8.8 & 0.57 & $-$1.5 & $-$0.1 & $-$0.60 & 261 \\
Sonnet 4.6 & 2026 & leaderboard & 0 & +8.5 & 0.70 & +1.5 & +0.2 & $-$0.13 & 165 \\
Gemini 3.5 Flash & 2020 & leaderboard & 0 & +18.6 & 0.52 & $-$11.2 & $-$0.9 & +0.36 & 73 \\
Gemini 3.5 Flash & 2020 & leaderboard & 1 & +9.3 & 0.53 & $-$15.9 & $-$1.4 & $-$0.13 & 93 \\
Gemini 3.5 Flash & 2020 & leaderboard & 2 & +16.7 & 0.62 & $-$7.7 & $-$0.7 & $-$0.01 & 82 \\
Gemini 3.5 Flash & 2020 & rerun & 0 & +10.6 & 0.48 & $-$12.7 & $-$1.2 & +0.14 & 94 \\
Gemini 3.5 Flash & 2022 & leaderboard & 0 & $-$8.9 & 0.75 & $-$5.6 & $-$0.7 & $-$0.54 & 85 \\
Gemini 3.5 Flash & 2022 & leaderboard & 1 & $-$5.2 & 0.74 & $-$2.0 & $-$0.3 & +0.20 & 95 \\
Gemini 3.5 Flash & 2022 & leaderboard & 2 & $-$5.1 & 0.70 & $-$2.4 & $-$0.3 & $-$0.16 & 89 \\
Gemini 3.5 Flash & 2022 & rerun & 0 & $-$6.8 & 0.75 & $-$3.5 & $-$0.4 & $-$0.35 & 92 \\
Gemini 3.5 Flash & 2026 & leaderboard & 0 & +9.4 & 0.75 & +2.0 & +0.2 & +0.86 & 70 \\
Gemini 3.5 Flash & 2026 & leaderboard & 1 & $-$0.6 & 0.80 & $-$9.6 & $-$1.1 & +0.16 & 114 \\
Gemini 3.5 Flash & 2026 & leaderboard & 2 & +1.7 & 0.80 & $-$5.9 & $-$0.6 & $-$0.17 & 111 \\
Gemini 3.5 Flash & 2026 & rerun & 0 & +0.8 & 0.76 & $-$5.5 & $-$0.5 & +0.30 & 116 \\
GPT-5.6 Luna & 2020 & leaderboard & 0 & +12.5 & 0.60 & +1.1 & +0.2 & $-$0.48 & 65 \\
GPT-5.6 Luna & 2020 & leaderboard & 1 & +20.7 & 0.76 & +1.7 & +0.4 & +0.01 & 68 \\
GPT-5.6 Luna & 2020 & leaderboard & 2 & +17.9 & 0.76 & +2.4 & +0.4 & $-$0.93 & 33 \\
GPT-5.6 Luna & 2020 & rerun & 0 & +15.5 & 0.73 & $-$1.8 & $-$0.2 & +0.30 & 62 \\
GPT-5.6 Luna & 2020 & rerun & 1 & +14.5 & 0.63 & $-$2.8 & $-$0.5 & $-$0.01 & 49 \\
GPT-5.6 Luna & 2020 & rerun & 2 & +5.5 & 0.78 & $-$10.8 & $-$1.2 & $-$0.15 & 59 \\
GPT-5.6 Luna & 2022 & leaderboard & 0 & $-$8.0 & 0.51 & +0.7 & +0.1 & $-$0.11 & 70 \\
GPT-5.6 Luna & 2022 & leaderboard & 1 & $-$9.4 & 0.48 & $-$0.9 & $-$0.2 & $-$0.15 & 92 \\
GPT-5.6 Luna & 2022 & leaderboard & 2 & $-$10.7 & 0.66 & $-$0.3 & $-$0.0 & $-$0.17 & 71 \\
GPT-5.6 Luna & 2022 & rerun & 0 & $-$8.2 & 0.52 & +1.2 & +0.3 & $-$0.16 & 73 \\
GPT-5.6 Luna & 2022 & rerun & 1 & $-$14.1 & 0.60 & $-$3.4 & $-$0.6 & +0.09 & 56 \\
GPT-5.6 Luna & 2022 & rerun & 2 & $-$7.8 & 0.60 & +1.7 & +0.4 & $-$0.02 & 78 \\
GPT-5.6 Luna & 2026 & leaderboard & 0 & +5.1 & 0.60 & $-$4.0 & $-$0.5 & $-$0.41 & 118 \\
GPT-5.6 Luna & 2026 & leaderboard & 1 & +11.4 & 0.70 & +4.3 & +0.8 & $-$0.59 & 93 \\
GPT-5.6 Luna & 2026 & leaderboard & 2 & +8.0 & 0.76 & +0.4 & +0.1 & $-$0.19 & 85 \\
GPT-5.6 Luna & 2026 & rerun & 0 & +7.5 & 0.76 & $-$2.9 & $-$0.6 & +0.02 & 65 \\
GPT-5.6 Luna & 2026 & rerun & 1 & +14.3 & 0.90 & +4.7 & +0.8 & $-$0.59 & 58 \\
GPT-5.6 Luna & 2026 & rerun & 2 & +8.3 & 0.68 & +1.3 & +0.2 & +0.16 & 89 \\
Llama 3.2 3B & 2020 & leaderboard & 0 & +14.8 & 0.23 & +2.9 & +0.6 & +2.47 & 3 \\
Llama 3.2 3B & 2020 & leaderboard & 1 & +15.0 & 0.23 & +3.3 & +0.7 & +3.76 & 4 \\
Llama 3.2 3B & 2020 & leaderboard & 2 & +16.2 & 0.23 & +3.4 & +0.7 & +3.45 & 2 \\
Llama 3.2 3B & 2022 & leaderboard & 0 & $-$7.6 & 0.25 & +0.3 & +0.1 & $-$0.47 & 13 \\
Llama 3.2 3B & 2022 & leaderboard & 1 & $-$4.4 & 0.22 & +1.6 & +0.6 & $-$0.12 & 14 \\
Llama 3.2 3B & 2022 & leaderboard & 2 & $-$14.4 & 0.52 & $-$3.1 & $-$0.5 & $-$0.19 & 15 \\
Llama 3.2 3B & 2026 & leaderboard & 0 & +10.6 & 0.30 & +5.6 & +0.7 & $-$2.23 & 2 \\
Llama 3.2 3B & 2026 & leaderboard & 1 & +6.3 & 0.58 & +0.1 & +0.0 & $-$1.32 & 7 \\
Llama 3.2 3B & 2026 & leaderboard & 2 & +4.9 & 0.30 & $-$1.0 & $-$0.1 & +0.10 & 11 \\
\bottomrule
\end{tabular}
\end{table*}

\section{Intraday Results for All 56 Stocks}
\label{app:latency}

\Cref{sec:latency} studies how fast prices took in the news for ten stocks. \Cref{tab:latencyall} repeats its hourly analysis (\cref{tab:latency}) for all 56 stocks of the benchmark, with \latAllEvents~events, \latAllEarnEvents~of them about earnings or guidance. The columns are the same as in \cref{tab:latency}.

\begin{table}[t]
\caption{The hourly analysis of \cref{tab:latency} for all 56 stocks: return over SPY in the direction of the news (\%) after new-fact headlines in 2026, with $t$-statistics in parentheses.}
\label{tab:latencyall}
\centering
\footnotesize
\setlength{\tabcolsep}{3.5pt}
\begin{tabular}{@{}lrrrrr@{}}
\toprule
News & Events & Whole & First hour & Rest & Next open \\
\midrule
\multicolumn{6}{@{}l}{\emph{Earnings and guidance}} \\
\quad Outside regular hours & 225 & +1.48 & +0.83 & +0.55 & +0.41 \\
 & & (4.1) & (3.8) & (1.97) & (1.6) \\
\quad Regular hours & 116 & $-$0.27 & $-$0.02 & $-$0.36 & $-$0.33 \\
 & & ($-$0.9) & ($-$0.2) & ($-$1.4) & ($-$1.7) \\
\quad All & 341 & +0.88 & +0.54 & +0.24 & +0.16 \\
 & & (3.2) & (3.5) & (1.2) & (0.9) \\
\multicolumn{6}{@{}l}{\emph{Other news}} \\
\quad Outside regular hours & 1,402 & +0.31 & +0.15 & +0.06 & +0.02 \\
 & & (3.9) & (4.8) & (0.8) & (0.2) \\
\quad Regular hours & 1,088 & +0.09 & +0.04 & $-$0.06 & $-$0.14 \\
 & & (1.1) & (1.5) & ($-$0.8) & ($-$2.7) \\
\quad All & 2,490 & +0.21 & +0.11 & +0.01 & $-$0.05 \\
 & & (3.5) & (4.7) & (0.1) & ($-$1.1) \\
\bottomrule
\end{tabular}
\end{table}

\end{document}